\documentclass[10pt,twocolumn,letterpaper]{article}

\usepackage[pagenumbers]{cvpr} 

\definecolor{cvprblue}{rgb}{0.21,0.49,0.74}
\usepackage[pagebackref,breaklinks,colorlinks,allcolors=cvprblue]{hyperref}

\usepackage{hyperref}       
\usepackage{url}            
\usepackage{booktabs}       
\usepackage{amsfonts}       
\usepackage{nicefrac}       
\usepackage{microtype}      
\usepackage{xcolor}         
\definecolor{aliceblue}{RGB}{240,248,255}
\usepackage{colortbl}
\usepackage{amsmath} 
\usepackage {amssymb}

\usepackage{graphicx}
\usepackage[T1]{fontenc}

\usepackage{tikz,graphics,color,float,epsf,caption,subcaption}
\usepackage{booktabs, 
            makecell, multirow, tabularx}
\usepackage{algorithm}
\usepackage{algorithmic}

\def\paperID{3312} 
\def\confName{CVPR}
\def\confYear{2025}

\title{VARGPT: Unified Understanding and Generation in a Visual Autoregressive Multimodal Large Language Model}

\author{Xianwei Zhuang~\thanks{Equal contribution}~,  Yuxin Xie~$^*$,  Yufan Deng~$^*$, Liming Liang, Jinghan Ru, Yuguo Yin, Yuexian Zou~\thanks{Corresponding author}\\
SECE of Peking University\\
{\tt\small xwzhuang@stu.pku.edu.cn} \\
\textcolor{red}{Project Page: \url{https://vargpt-1.github.io/}}
}

\begin{document}
\maketitle

\begin{abstract}

We present VARGPT, a novel multimodal large language model (MLLM) that unifies visual understanding and generation within a single autoregressive framework. VARGPT employs a next-token prediction paradigm for visual understanding and a next-scale prediction paradigm for visual autoregressive generation. VARGPT innovatively extends the LLaVA architecture, achieving efficient scale-wise autoregressive visual generation within MLLMs while seamlessly accommodating mixed-modal input and output within a single model framework.
Our VARGPT undergoes a three-stage unified training process on specially curated datasets, comprising a pre-training phase and two mixed visual instruction-tuning phases.
The unified training strategy are designed to achieve alignment between visual and textual features, enhance instruction following for both understanding and generation, and improve visual generation quality, respectively.
Despite its LLAVA-based architecture for multimodel understanding, VARGPT significantly outperforms LLaVA-1.5 across various vision-centric benchmarks, such as visual question-answering and reasoning tasks. Notably, VARGPT naturally supports capabilities in autoregressive visual generation and instruction-to-image synthesis, showcasing its versatility in both visual understanding and generation tasks.

\end{abstract}

\section{Introduction}
\label{sec:intro}

\begin{figure}[t]
   \centering
    \includegraphics[width=1\linewidth]{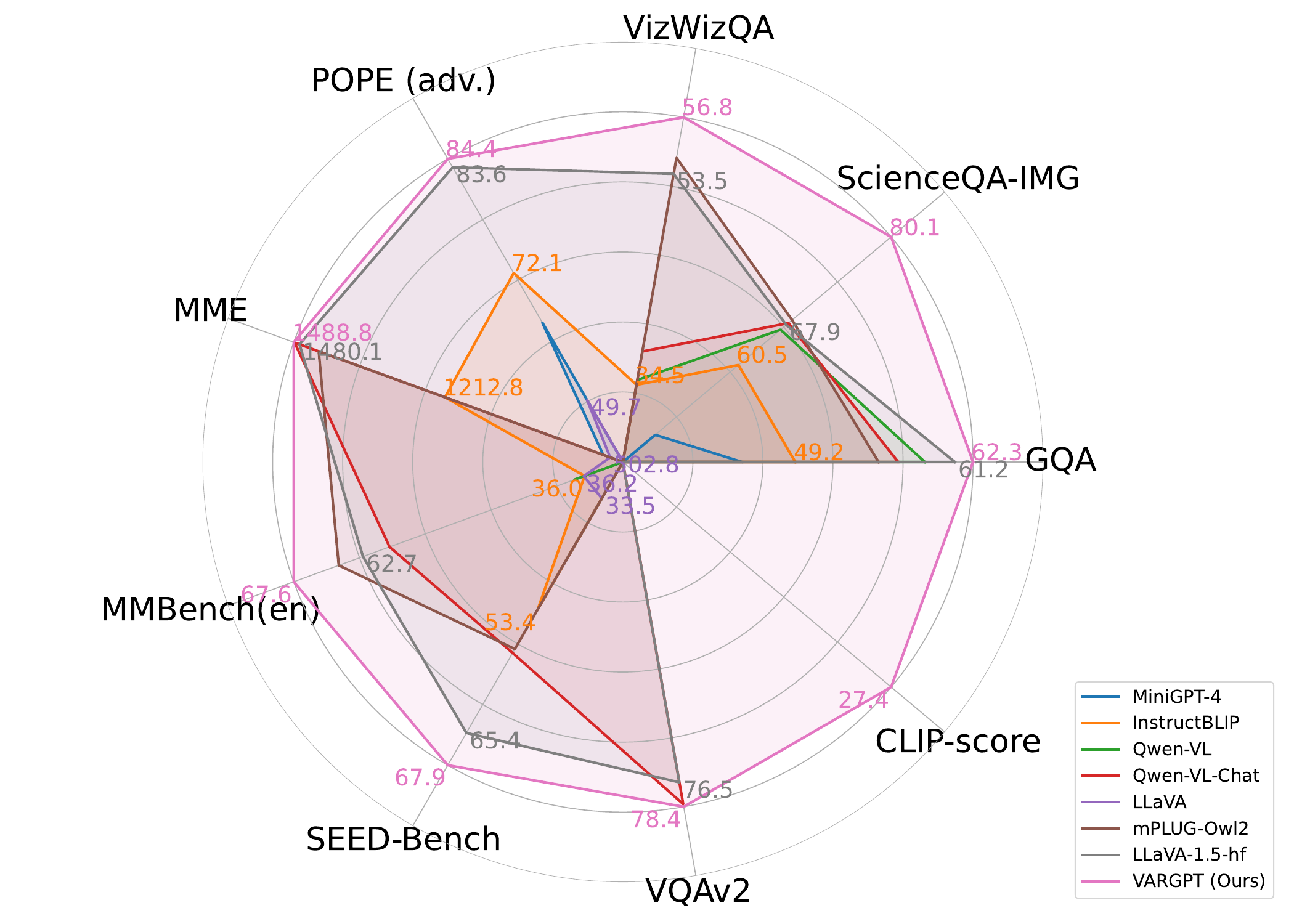}
    \caption{A comparative analysis of various MLLMs across multiple visual comprehension 
 and generation benchmarks is presented. The CLIP-scores is employed as a text-to-image visual generation metric, while the remaining metrics are derived from standard visual question-answering benchmarks and multi-modal comprehension benchmarks. Notably, our VARGPT model demonstrates significant superiority over the compared baselines across all comprehension benchmarks. Furthermore, it exhibits exceptional instruction-to-image generation capabilities, thus enhancing its versatility and applicability in diverse visual-linguistic tasks.}
    \label{fig:intro}
 \end{figure}

\begin{figure*}[t]
   \centering

    \includegraphics[width=1.0\linewidth]{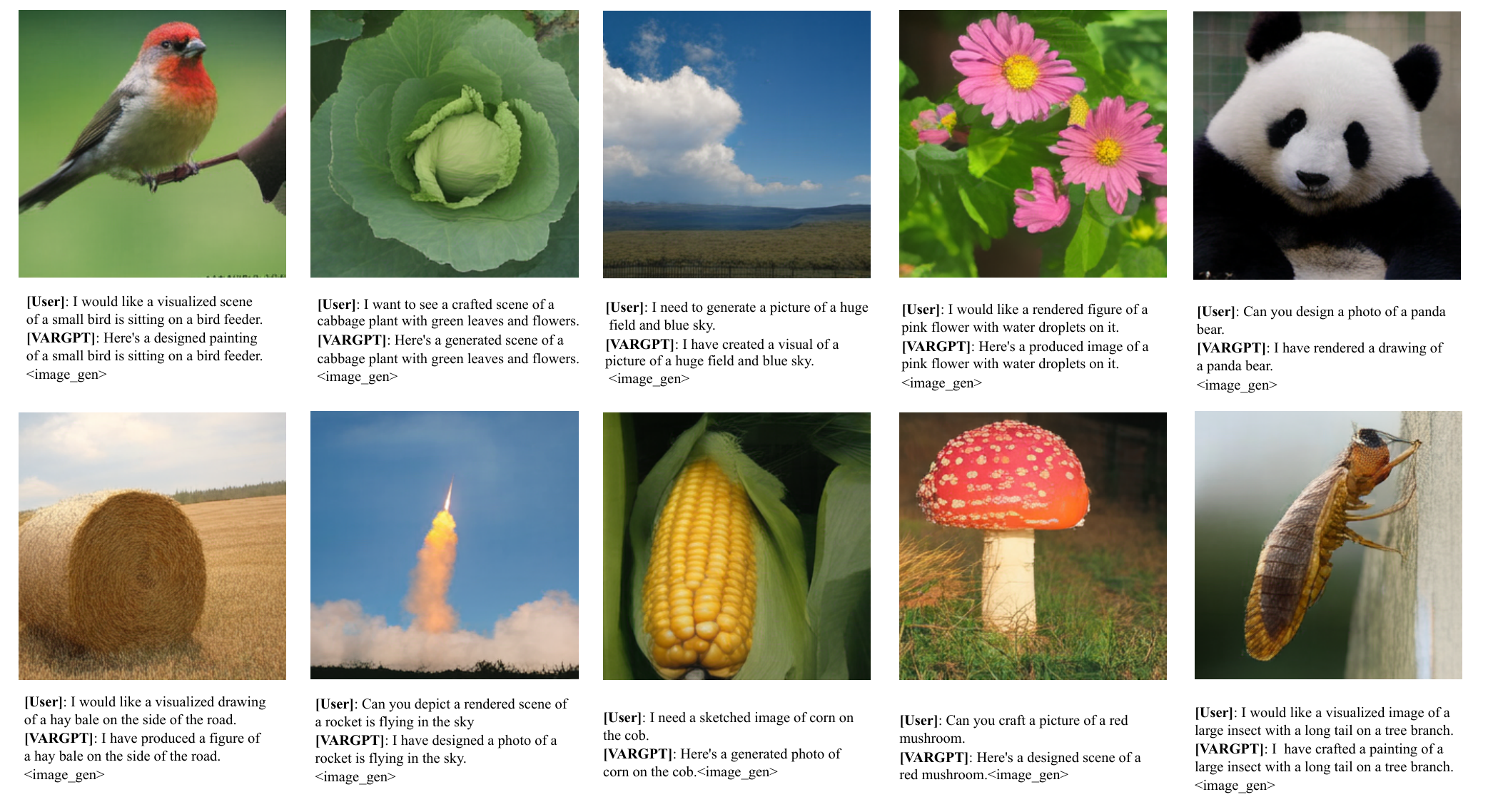}
    \captionof{figure}{
    Some generated 256$\times$256 samples by VARGPT trained on ImageNet~\cite{deng2009imagenet}. VARGPT supports text-and-image instructions from user and outputs both text-and-image mixed modal data simultaneously.} 
    \label{fig:abs_generation}
 \end{figure*}

\begin{figure*}[t]
  \centering
   \includegraphics[width=1\linewidth]{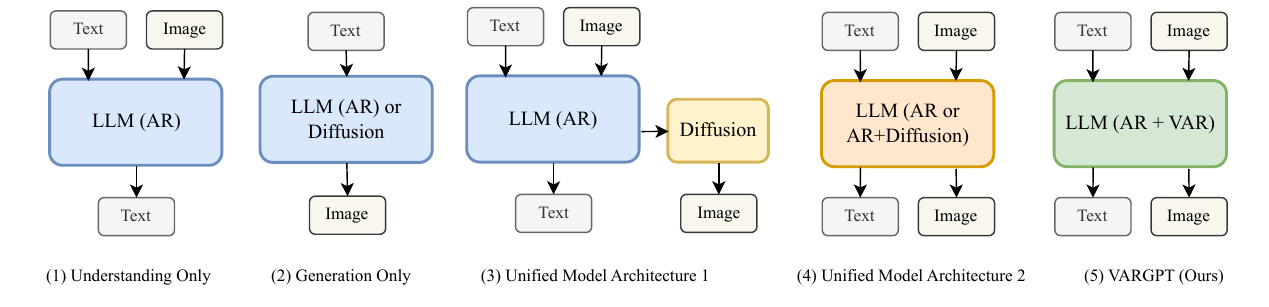}

   \caption{Comparison of different model architectures, where, `AR' denotes autoregressive, while `VAR' signifies visual autoregressive. We present a comparative analysis of architectures designed for comprehension-only tasks, generation-only tasks, and unified comprehension and generation, alongside our proposed VARGPT model. Our VARGPT is conceptualized as a purely autoregressive multimodel model, achieving visual comprehension through next-token prediction and visual generation through next-scale prediction paradigms.} 
   \label{fig:vargpt_arch}
\end{figure*}

In recent years, multimodal artificial intelligence has achieved significant breakthroughs in the two core domains of understanding and generation. Multimodal Large Language Models (MLLMs)~\cite{Liu2023VisualIT, Chen2023MiniGPTv2LL, Ye2023mPLUGOwl2RM, dai2023instructblipgeneralpurposevisionlanguagemodels, bai2023qwen} have demonstrated exceptional capabilities in understanding multimodal data by leveraging the strong generality of LLMs~\cite{touvron2023llama, touvron2023llama-2, bai2023qwen}. Concurrently, Denoising Diffusion Probabilistic Models (DDPMs)~\cite{ho2020denoising,nichol2021improved} have brought substantial advancements to the field of image generation, achieving superior performance in text-to-visual modality generation. Moreover, inspired by the advantageous properties of autoregressive LLMs, such as the scaling law~\cite{kaplan2020scaling, henighan2020scaling}, numerous works have explored autoregressive visual generation by predicting the next token or next scale, exemplified by Emu3~\cite{wang2024emu3}, VAR~\cite{VAR}, LlamaGen~\cite{sun2024autoregressive}, HART~\cite{tang2024hart}, and Infinity~\cite{Infinity}, yielding notable results.
Given these achievements in visual understanding and generation, recent studies have been exploring unified models capable of handling understanding and generation, consequently designing various unified architectures to achieve this objective (as illustrated in Figure~\ref{fig:vargpt_arch}).
Recent works~\cite{wu2023next,ge2024seed,xie2024showo} have attempted to assemble models from these two distinct domains (e.g., LLMs and DDPMs) to form a unified system capable of processing multimodal understanding and generation (as shown in Figure~\ref{fig:vargpt_arch}.(3)). For instance, NExT-GPT~\cite{wu2023next} and SEED-X~\cite{ge2024seed} may rely on pre-trained diffusion models for image generation. Furthermore, LWM~\cite{liu2023world}, Chameleon~\cite{Chameleon_Team_Chameleon_Mixed-Modal_Early-Fusion_2024} and Janus~\cite{wu2024janusdecouplingvisualencoding} have explored purely next-token prediction unified models (as depicted in Figure~\ref{fig:vargpt_arch}.(4)), while Dual Diffusion~\cite{li2024dualdiffusionunifiedimage} investigated the use of two diffusion models for understanding and generation.
TokenFlow~\cite{qu2024tokenflowunifiedimagetokenizer} has explored a unified image tokenizer, but the generation and understanding models are separate. Show-o~\cite{xie2024showo} proposed combining autoregressive and diffusion model paradigms within a single Transformer (as illustrated in Figure~\ref{fig:vargpt_arch}.(4)). Liquid~\cite{wu2024liquid} learns images and text embeddings within the same space and implementing autoregressive visual understanding and generation using the paradigm of predicting the next token.

In this work, we endeavor to unify visual generation and understanding within a visual autoregressive MLLM, naturally supporting mixed-modal input and output. Different from all existing unified model, we propose modeling understanding and generation as two distinct paradigms within a unified model: \textbf{predicting the next token for visual understanding and predicting the next scale for visual generation}, respectively, and train a novel unified model called VARGPT. Specifically:

(1) In terms of \textbf{model architecture}, the core structure of our VARGPT draws inspiration from LLaVA-1.5-7B, while we additionally incorporate a visual decoder and two extra visual feature projectors for visual generation. These projectors are used for the mutual mapping between the generated visual and textual features.  Our VARGPT employs an autoregressive approach to predict next textual tokens for visual understanding and question answering. When a special token for visual generation is predicted, the model autoregressively predicts the next-scale tokens and obtains the final output image through the visual decoder. The proposed architecture enables VARGPT to achieve unified understanding and generation within a visual autoregressive MLLM.

(2) Regarding the \textbf{training method}, we employ unified instruction-tuning to learn visual understanding and visual generation. Specifically, we extend the instruction tuning to visual generation through construct visual token prediction as the instruction-following formats and combine the constructed visual generation instruction datasets with the multi-turn dialogue instruction datasets from LLaVA-1.5~\cite{llava} for mixed training. Through proposed unified instruction tuning, we simultaneously endow MLLMs with understanding and generation capabilities. We divide the training process into three stages, including a pre-training stage and two instruction fine-tuning stages. In the first stage of pre-training, the model learns feature mapping between textual and visual space. 
In the second and third stages of instruction tuning, VARGPT enhances its capabilities in visual question answering and instruction-to-image generation, respectively.

(3) Concerning \textbf{training datasets}, to train the model efficiently, we have constructed and collected 1.28M pre-training data for the first stage, 1.18M mixed visual understanding and generation instruction fine-tuning data for the second stage, and 1.4M visual generation instruction-tuning data for the third stage.
Through a unified instruction-following format, we unify the training of understanding and generation within mixed visual instructions tuning.

Extensive experiments demonstrate that our VARGPT can achieve significant visual understanding capabilities (as shown in Figure~\ref{fig:intro}) and endow MLLMs with visual generation abilities, naturally supporting mixed-modal input and output (as illustrated in Figure~\ref{fig:abs_generation}). To the best of our knowledge, VARGPT is the first unified model to support predicting the next token for understanding tasks and predicting the next scale for generation tasks, while surpassing numerous MLLMs and unified models of comparable scale in terms of comprehension capabilities.

In summary, our main contributions are threefold:
\begin{itemize}
\item We explore a novel unified architecture for autoregressive visual understanding and generation, which employs a next-token paradigm for visual understanding and a next-scale paradigm for visual generation.
\item We develop VARGPT, a unified model supporting mixed-modal input and output, through the novel architecture, a proposed three-stage training strategy, and a unified instruction-tuning dataset of 3.86M samples.
\item Extensive experiments and evaluations demonstrate that VARGPT achieves competitive performance on numerous multimodal understanding and vision-centric benchmarks, while simultaneously exhibiting exceptional capabilities in autoregressive text-to-image synthesis.
\end{itemize}

\begin{figure*}[t]
  \centering
   \includegraphics[width=0.95\linewidth]{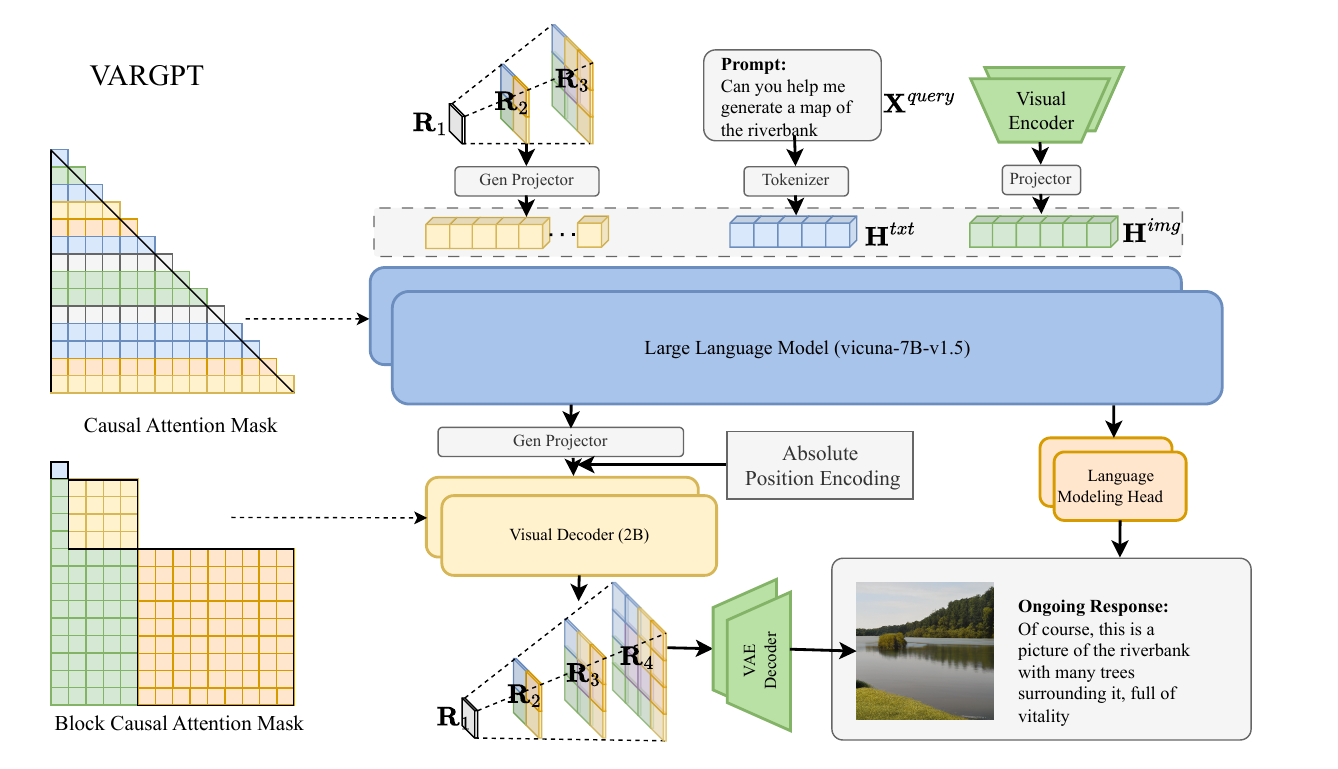}

   \caption{The illustration of the proposed VARGPT framework, which consists of (1) a large language model, visual encoder and a understanding projector for visual understanding; (2) a visual decoder and dual generation projectors for visual generation. VARGPT employs the causal attention in the LLM backbone, while utilizing block causal attention in the visual decoder.}
   \label{fig:main}
\end{figure*}

\section{Related Work}
\label{sec:related}
\textbf{Visual Generation.}
Diffusion models~\cite{ho2020denoisingdiffusionprobabilisticmodels, song2020generativemodelingestimatinggradients, song2022denoisingdiffusionimplicitmodels} frame image generation as a reverse diffusion process from noise to image. Advances in diffusion models have primarily focused on sampling methods~\cite{bao2022analyticdpmanalyticestimateoptimal,lu2022dpmsolverfastodesolver} and architectural design~\cite{ho2021cascadeddiffusionmodelshigh,podell2023sdxlimprovinglatentdiffusion}, leading to impressive models such as~\cite{ma2023unifiedmultimodallatentdiffusion,podell2023sdxlimprovinglatentdiffusion}. Against the backdrop of significant progress in diffusion models, flow-based~\cite{albergo2023buildingnormalizingflowsstochastic} generative models have emerged as a simplified framework, driving the development of advanced visual generation models. Autoregressive models~\cite{esser2021tamingtransformershighresolutionimage,yu2022scalingautoregressivemodelscontentrich} employ GPT-style~\cite{radford2018improving} techniques to predict the next token in a sequence. Works like~\cite{ding2021cogviewmasteringtexttoimagegeneration, ramesh2021zeroshottexttoimagegeneration, wang2024emu3, sun2024autoregressive, sun2024autoregressivemodelbeatsdiffusion,fan2024fluid} utilize visual tokenizers similar to VQGAN~\cite{lee2022autoregressiveimagegenerationusing} to convert images into discrete tokens, enabling the tokenization of visual data and adopting prediction methods similar to GPT-style approaches. Recently, another class of autoregressive models, e.g., VAR~\cite{VAR}, HART~\cite{tang2024hart} and Infinity~\cite{Infinity} based on predicting the next scale has garnered attention and has been verified to potentially possess properties consistent with scaling laws~\cite{kaplan2020scaling, henighan2020scaling}. In this work, our unified autoregressive framework accomplishes image generation tasks through the paradigm of predicting the next scale.

\textbf{Multimodel Large Language Model.} The advancements in LLMs~\cite{touvron2023llama, touvron2023llama-2} have propelled the development of MLLMs. MLLMs utilize pre-trained LLMs as text decoders, integrating text and images by connecting them with visual encoders through connectors~\cite{li2023blip,lecun2015deep}. LLaVA~\cite{llava} fine-tunes the model using data from various tasks (e.g., visual question answering and image caption) in an instructional format, enabling the model to understand novel instructions, and generalize to unseen tasks. The LLaVA-1.5~\cite{liu2024improved} and LLaVA-NeXT~\cite{liu2024llavanext,zhang2024llavanext-video,li2024llavanext-strong,li2024llavanext-ablations,li2024llavanext-interleave} series have further enhanced visual understanding performance through more diverse and higher-quality datasets. With architectural optimizations, innovative training paradigms, and the introduction of diverse data, a series of advanced MLLMs have emerged, such as Qwen-VL~\cite{bai2023qwen}, mPLUG-Owl2~\cite{ye2024mplug}, InternVL~\cite{chen2024internvl}, InstructBLIP~\cite{dai2023instructblipgeneralpurposevisionlanguagemodels}.

\textbf{Unified Models For Visual Understanding and Generation. }
In recent years, researchers have been dedicated to unifying understanding and generation capabilities within a model~\cite{ye2024xvilacrossmodalityalignmentlarge,dong2024dreamllmsynergisticmultimodalcomprehension,tang2023anytoanygenerationcomposablediffusion}. Most existing approaches~\cite{sun2024emugenerativepretrainingmultimodality,wu2023next,ge2023makingllamadrawseed} attempt to integrate pre-trained diffusion models with existing systems. However, these systems essentially treat diffusion models as external tools rather than incorporating them as intrinsic generative capabilities of MLLMs. Show-o~\cite{xie2024showosingletransformerunify} by combining autoregressive and (discrete) diffusion modeling, can adaptively handle inputs and outputs of various mixed modalities. \citet{li2024dualdiffusionunifiedimage} employs a cross-modal maximum likelihood estimation framework, significantly improving upon existing diffusion-based multimodal models. ~\cite{geminiteam2024geminifamilyhighlycapable,bai2023sequentialmodelingenablesscalable} have explored integrating image generation into large language models (LLMs) using autoregressive methods, achieving remarkable results. For instance, LWM~\cite{liu2023world} and Chameleon~\cite{chameleonteam2024chameleonmixedmodalearlyfusionfoundation} utilizes a VQ tokenizer~\cite{esser2021tamingtransformershighresolutionimage, VAR} to encode images, enabling simultaneous support for multimodal understanding and generation. Janus~\cite{wu2024janusdecouplingvisualencoding} further enhances model flexibility and performance by decoupling visual encoding into separate pathways, while Dual Diffusion~\cite{li2024dualdiffusionunifiedimage} investigated the use of two diffusion models for understanding and generation. Liquid~\cite{wu2024liquid} learns images and text embeddings within the same space and implementing autoregressive visual understanding and generation using the paradigm of predicting the next token.
Different from all existing unified model, we propose modeling understanding and generation as two distinct paradigms within a unified model: {predicting the next token for visual understanding and predicting the next scale for visual generation}.

\section{Methodology}
\label{sec:methods}

\begin{figure*}[t]
  \centering
   \includegraphics[width=1\linewidth]{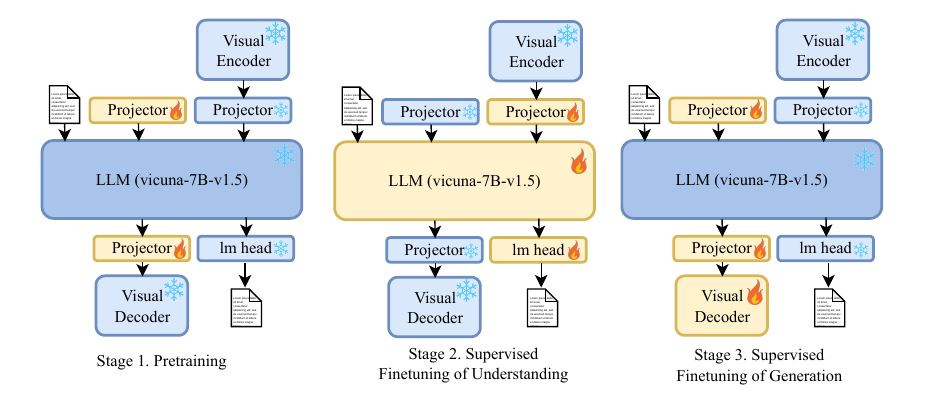}

   \caption{The three training stages of the VARGPT, including stage-1 pretraining, stage-2 and stage-3 instruction fine-tuning.}
   \label{fig:vargpt_training}
\end{figure*}

\subsection{Model Architecture}
Our VARGPT unifies visual understanding and generation, which architecture is shown in the Figure~\ref{fig:main}. Our architecture follows the paradigm of predicting the next token for understanding and question answering, and adheres to the paradigm of predicting the next scale for image generation.

\textbf{Visual understanding via next-token prediction}~
For visual understanding, our model architecture references the structure of LLaVA-1.5~\cite{liu2024improved}, using Vicuna-7B-v1.5~\cite{zheng2023judgingllmasajudgemtbenchchatbot} as the LLM $\theta$, and employing CLIP's~\cite{radford2021learningtransferablevisualmodels} visual encoder (ViT/14) as the visual encoder along with a two-layer linear network as the projector.
Initially, the image $\mathbf{X}^{img}$ for visual understanding undergoes processing through the vision encoder to produce embeddings $\mathbf{H}^{img}$, which are then modified by the interface (e.g., linear layer) to align with the textual embeddings $\mathbf{H}^{txt}$ obtained by query $\mathbf{X}^{query}$. The combined data serves as input to the LLM, which autoregressively generates the textual output $\mathbf{Y}^{txt}$ as:
\begin{equation}
    \mathbf{Y}^{txt}_t \sim p_{\theta}(\mathbf{Y}^{txt}_t \mid \mathbf{X}^{img}, \mathbf{X}^{query}, \mathbf{Y}^{txt}_{<t}),
\end{equation}
where $\mathbf{Y}^{txt}_t$ represents the $t$-th token of $\mathbf{Y}^{txt}$, and $\mathbf{Y}^{txt}_{<t}$ refers to the sequence of tokens generated prior to the $t$-th step.
To maintain the causal attention property of LLM, we apply a causal attention mask to all input LLM tokens, including those used for generating images.

\textbf{Visual generation via next-scale prediction}~
For visual generation, we follow most of the settings in VAR~\cite{VAR} and adopt multi-scale image tokenizer for visual token encoding and decoding. We construct two image generation projectors to transform the visual features used for generation at the input and output of the LLM. Additionally, we construct an extra 2B visual decoder $\phi$ with 30 layers of Transformers to decode visual features, which can to some extent avoid conflicts between knowledge in the text decoder and image generation knowledge. The image features obtained through the visual decoder will be further decoded through the multi-scale VAE decoder to produce usable images. Unlike the text decoder (i.e., the LLM), the visual decoder uses attention that follows the block causal attention in VAR~\cite{VAR} to support predicting tokens for the next scale. Furthermore, before feeding the features for visual generation into the visual decoder, we add absolute positional encoding to further distinguish the positional information of visual tokens.

Formally, we define the multi-scale feature maps of a image as $(\mathbf{R}_1, \mathbf{R}_2, \cdots, \mathbf{R}_K )$ obtained by the multi-scale tokenizer. Therefore, the image tokens for the next scale will be generated in an autoregressive manner:
\begin{equation}
    \mathbf{R}_t \sim  p_{\{\theta, \phi\}} (\mathbf{R}_t \mid \mathbf{X}^{query}, \mathbf{R}_{<t}).
\end{equation}

\textbf{Prompt template for mixed-modal generation}~
To differentiate between tokens designated for textual generation and those for image synthesis, we design some special token markers. Specifically, we utilize \texttt{<image\_gen>} for image generation token position filling, \texttt{<image\_gen\_start>} to denote the initiation of image generation tokens, and \texttt{<image\_gen\_end>} to signify the end of generation.  Upon the generation of the \texttt{<image\_gen\_start>} token by VARGPT, the features associated with the \texttt{<image\_gen>} token are processed through a projector and subsequently fed into the visual decoder to derive features requisite for image generation. In the context of visual comprehension tasks, we employ the \texttt{<image>} token as a representation of the input image.
We summarize the prompt template using by our VARGPT in Appendix~\ref{sec:appendix-model-details}.

\textbf{Classifier-free guidance (CFG)}~
CFG significantly enhances the capability of generative diffusion models to produce samples of exceptionally high fidelity. This approach integrates conditional generative models with the distribution estimation of unconditional models trained concurrently, thereby improving the overall quality of generation.
Inspired by DALL-E 2~\cite{ramesh2022hierarchicaltextconditionalimagegeneration}, VAR~\cite{VAR} and VAR-CLIP~\cite{zhang2024varcliptexttoimagegeneratorvisual}, we employ Gaussian noise as input to simulate unconditional generation. Subsequently, we derive the final distribution of visual outputs by subtracting the probability of unconditional generation from the logits distribution of conditional generation. 
We provide more details in Appendix~\ref{sec:appendix-model-details}.

\subsection{Training}

For VARGPT model training, we propose a one-stage pretrainined procedure and a two-stage instruction-tuning procedure, as shown in Figure~\ref{fig:vargpt_training}.

\subsubsection{Stage-1: Pretraining}
We use images from ImageNet~\cite{deng2009imagenet} as the image source to construct training data for pre-training the two projectors used for image generation. We structure the pre-training data into 1.28M single-round dialogue data (for specific data construction, please refer to our Sec.~\ref{sec:dataset}). The main purpose of this pre-training stage is to train the projectors to preliminarily align image generation features with text features. During pre-training, we freeze all parameters except for the two projetors for image generation, as shown in Figure~\ref{fig:vargpt_training}.

\subsubsection{Stage-2: SFT for Visual Understanding}
In the second stage, we unfreeze the language model and the projectors for visual encoder feature output, and use our curated multi-turn dialogue and understanding datasets for training. The main purpose of this stage is to ensure that VARGPT maintains excellent multi-turn dialogue, visual understanding  and question-answering capabilities. Additionally, in this stage, we introduce 5K samples from our constructed Imagenet-Instruct dataset, enabling VARGPT to distinguish between visual understanding and visual generation tasks. When users input generation instructions, VARGPT can accurately respond by outputting the special token \texttt{<image\_gen\_start>} to begin autoregressive visual generation.
The composition of our dataset for the stage-2 training can be referred to in Sec.~\ref{sec:dataset}.

\subsubsection{Stage-3: SFT for Visual Generation}
Compared to the second stage, the third stage primarily aims to improve VARGPT's ability to instruction-to-image through supervised fine-tuning.
In this stage, we unfreeze the visual decoder and the two projectors used for visual generation, while freezing other parameters to perform SFT, as shown in Figure~\ref{fig:vargpt_training} Stage-3.
The training data for the third stage includes 1400K instruction pairs constructed from ImageNet (refer to Sec~\ref{sec:dataset} for details).

\section{Unified Instruction-following Data}
\label{sec:dataset}

\begin{figure}[t]
  \centering
   \includegraphics[width=1\linewidth]{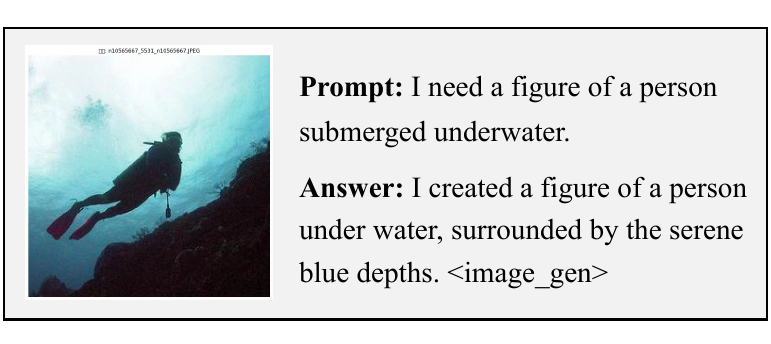}
   \caption{A sample from the ImageNet-Instruct-130K image generation instruction-following dataset, whose caption is a person
submerged underwater. \textless image\_gen \textgreater represents the special token used for image generation token position filling.}
   \label{fig:instruction-following-sample}
\end{figure}

\begin{figure*}[t]
  \centering
   \includegraphics[width=1\linewidth]{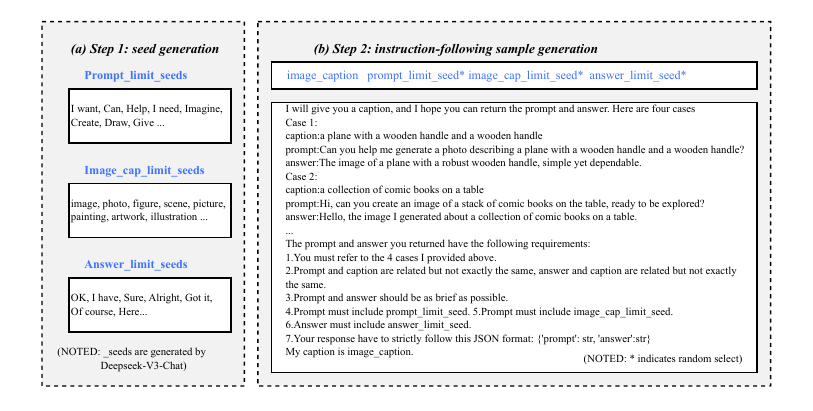}

   \caption{The illustration of the proposed image generation instruction-following sample generation, which consists of (a) seed generation: we utilize LLM to generate seeds for constraining the creation of the instruction-following dataset; (b) instruction-following sample generation: prompt template for the instruction-following dataset.}
   \label{fig:instruction-following-generate}
\end{figure*}

\begin{figure*}
    \centering
    \begin{subfigure}{0.47\linewidth}
      \includegraphics[width=1\linewidth]{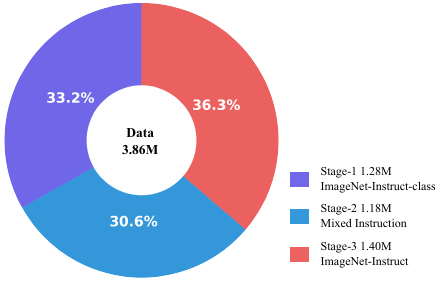}
      \caption{Data proportion in the three training stages.}
      \label{fig:PieChart2}
    \end{subfigure}
    \hfill
    \begin{subfigure}{0.47\linewidth}
      \includegraphics[width=1\linewidth]{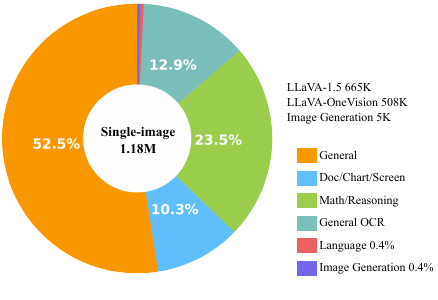}
      \caption{Data distribution of the mixed instruction fine-tuning dataset.}
      \label{fig:PieChart}
    \end{subfigure}
    \caption{We present the data distribution we constructed and collected, encompassing: (a) the proportional breakdown of data across the three training stages; and (b) the distribution of mixed instruction data employed during the instruction fine-tuning phase in the second stage. Our composite dataset for stage-2 training is derived from LLaVA-1.5, LLaVA-OneVision, and ImageNet-Instruct-130K.}
        \label{fig:PieChart-exp}
  \end{figure*}

\label{sec:data}
In this section, we delineate the origins of the training datasets utilized across three distinct stages and the proportional representation of diverse data types within them. Notably, we introduce image generation instruction-following datasets, shown in the Figure~\ref{fig:PieChart2}., elaborating on its provenance and the methodology underpinning its generation through the application of LLMs. 
Through this approach, we unify the training methods for visual understanding and generation into visual instruction fine-tuning.

\subsection{Generation Instruction-following Dataset}
We have constructed two image generation instruction-following datasets: ImageNet-Instruct-130K,  and ImageNet-Instruct-1270K. To illustrate, let us take the construction of ImageNet-Instruct-130K as an example. A sample from the dataset is illustrated in the Figure~\ref{fig:instruction-following-sample}.

\textbf{ImageNet-1K-VL-Enriched}~We employ the ImageNet-1K-VL-Enriched dataset~\cite{huggingface_dataset_imagenet-1k-vl-enriched} as our foundational dataset. ImageNet-1K-VL-Enriched is an enriched version of the ImageNet dataset, where image captions are generated using the BLIP2~\cite{li2023blip2bootstrappinglanguageimagepretraining} captioning model. 

\textbf{Constructing ImageNet-Instruct-130K through Deepseek-LLM}~To construct the question-and-answer format for the instruction fine-tuning dataset, we utilized Deepseek-V3 Chat LLM~\cite{deepseekai2024deepseekv3technicalreport} (hereafter referred to as LLM) to generate seed formats for prompts and answers (Prompt\_limit\_seeds and Answer\_limit\_seeds). As illustrated in Figure~\ref{fig:instruction-following-generate}(a), Prompt\_limit\_seeds effectively simulate user requests, while Answer\_limit\_seeds emulate the dialogue between the VLLM and the user. We randomly selected prompt\_limit\_seed, image\_cap\_limit\_seed, and answer\_limit\_seed from the seed pool to serve as elements in the LLM invocation template.

\textbf{LLM invocation template}~We randomly selected 4 image-caption samples from the base dataset as 4-shot examples to guide the large model in generating corresponding dialogue samples. As shown in Figure~\ref{fig:instruction-following-generate}(b), we added relevant constraints to the generated prompts and answers to ensure that the outputs are as compliant and diverse as possible.
We randomly sampled 130K image-caption data samples, resulting in the creation of 130K samples for the ImageNet~\cite{deng2009imagenet} image generation instruction fine-tuning dataset, which we have named ImageNet-Instruct-130K. We provide more details on the construction of dataset in Appendix~\ref{sec:appendix_data_construction}.

\begin{table*}[htbp]
  \centering
    \begin{tabular}{c|cc|ccccccc}
    \toprule
    \multirow{2}[2]{*}{\textbf{Methods}} & \multirow{2}[2]{*}{\textbf{LLM}} & \multicolumn{1}{c|}{\multirow{2}[2]{*}{\textbf{Gen.}}} & \multirow{2}[2]{*}{\textbf{MMBench}} & \multirow{2}[2]{*}{\textbf{SEED }} & \multirow{2}[2]{*}{\textbf{MMMU}} & \multicolumn{3}{c}{\textbf{POPE (acc.)}} & \multirow{2}[2]{*}{\textbf{MME }} \\
          &       &       &       &       &       & \textbf{adv} & \textbf{pop} & \textbf{rand} &  \\
    \midrule
    Chameleon (7B) & from scratch &  $\checkmark$     & 31.1  & 30.6  & 25.4  &       &       &       & 170 \\
    SEEDLLaMA (7B) & Vicuna-7B &  $\checkmark$     & 45.8  & 51.5  & -     & -     & -     & -     & - \\
    Show-o (1.5B) & Phi-1.5-1.3B &  $\checkmark$     & -     & -     & 25.1  & -     & -     & -     & 948.4 \\
    VILA-U(7B) & LLaMA-2-7B  &  $\checkmark$     & -     & 59.0  & -     & -     & -     & -     & 1401.8 \\
    Liquid (7B) & Gemma-7B &  $\checkmark$     & -     & -  & -     & -     & -     & 81.1     & 1119.3 \\
    \midrule
    MiniGPT-4 (7B) & Vicuna-7B &  $\times$     & 23.0  & 47.4  & \multirow{6}[1]{*}{-} & 65.17 & 69.73 & 79.67 & 581.7 \\
    InstructBLIP (8B) & Vicuna-7B & $\times$     & 36.0  & 53.4  &       & 72.10     & 82.77     & \textbf{88.57}     & 1212.8 \\
    Qwen-VL (7B) & Qwen-7B & $\times$     & 38.2  & -     &       & -     & -     & -     & - \\
    Qwen-VL-Chat (7B) & Qwen-7B & $\times$     & 60.6  & 58.2  &       & -     & -     & -     & 1487.5 \\
    LLaVA (7B) & Vicuna-7B & $\times$     & 36.2  & 33.5  &       & 49.70 & 49.87 & 50.37 & 502.8 \\
    mPLUG-Owl2 (7B)  & LLaMA-2-7B  & $\times$     & 64.5  & 57.8  &       & -     & -     & -     & 1450.2 \\
    LLaVA-1.5-hf (7B) & Vicuna-7B & $\times$     & 62.7  & 65.4  & 35.24 & 83.60  & 85.77 & 86.97 & 1480.1 \\
    \midrule
    \rowcolor{aliceblue} \textbf{VARGPT(7B+2B)} & \textbf{Vicuna-7B} &  $\checkmark$     & \textbf{67.6} & \textbf{67.9} & \textbf{36.44} & \textbf{84.40 } & \textbf{85.90 } & {87.37} & \textbf{1488.8} \\
    \bottomrule
    \end{tabular}%
  \caption{Zero-shot multi-modal evaluation on multi-modal benchmarks including MMMU, MME, MMBench, SEEDBench, and POPE (including different settings \textit{random}, \textit{popular} and \textit{adversarial} ). The overall scores are reported for evaluation and we report test results for MMBench. \textbf{Gen} represents whether the method supports image generation capability. VARGPT achieves the best overall
performance.}
  \label{tab:main-1}%
\end{table*}%
\begin{table}[htbp]
  \centering
    \begin{tabular}{l|ccc}
    \toprule
    \textbf{Hyperparameter} & \textbf{Stage 1} & \textbf{Stage 2} & \textbf{Stage 3} \\
    \midrule
    batch size & 1024  & 256   & 1024 \\
    lr    & 1e-3  & 5e-5  & 5e-5 \\
    lr schedule & \multicolumn{3}{c}{Cosine} \\
    lr warmup ratio & \multicolumn{3}{c}{0.1} \\
    weight decay & \multicolumn{3}{c}{0} \\
    epoch & 1     & 1     & 12 \\
    optimizer & \multicolumn{3}{c}{AdamW} \\
    DeepSpeed stage & 3     & 2     & 3 \\
    \bottomrule
    \end{tabular}%
  \caption{Main hyperparameters and training settings for training our VARGPT at various stages.}
  \label{tab:setting}%
\end{table}%

\subsection{Data Composition in Three Training Stages}
\textbf{Stage-1.}~The dataset ImageNet-Instruct-class with 1.28 million single-round dialogue samples for stage-1 pretraining is derived from ImageNet, focusing on learning the correspondence between categories and images. Assuming the category is `fish', the format is as follows:
\{`prompt': `Please generate an image of a fish for me.', `answer': `The generated image of a fish is as follows \textless image\textgreater\}.

\textbf{Stage-2.}~Our mixed instruction fine-tuning dataset in stage-2 is sourced from LLaVA-1.5~\cite{liu2024improvedbaselinesvisualinstruction}, LLaVA-OneVision~\cite{li2024llavaonevisioneasyvisualtask}, and ImageNet-Instruct-130K. Components are as shown in the Figure~\ref{fig:PieChart}.
\begin{itemize}
    \item LLaVA-1.5-665K. The instruction-following dataset of LLaVA-1.5 comprises VQA~\cite{goyal2017vqav2,hudson2019gqa,okvqa,schwenk2022okvqa}, OCR~\cite{mishra2019ocrvqa,sidorov2020textcaps}, region-level VQA ~\cite{kazemzadeh2014referitgame,mao2016generation,krishna2017visual}, visual conversation~\cite{llava}, and language conversation~\cite{sharegpt} data. We incorporate all 665K instruction-following samples into the training for stage-2.
    \item LLaVA-OneVision. The visual instruction tuning data of LLaVA-OneVision integrates data from LLaVA-1.5 and subsequent multiple LLaVA-NeXT versions~\cite{liu2024llavanext,zhang2024llavanext-video,li2024llavanext-strong,li2024llavanext-ablations,li2024llavanext-interleave}, as well as accumulates open-source datasets from the internet, with specific formatted prompts set to integrate data and avoid conflicts. This ultimately forms a high-quality single-image dataset of 3.2M. After removing samples from the K12 Printing subset, we randomly sampled 508K samples from this dataset for inclusion in the stage-2 training. (It is worth noting that we only sampled 5K pure text question answer pairs.)
    \item ImageNet-Instruct-130K. We randomly sampled 5K samples from the ImageNet-Instruct-130K dataset for inclusion in the training of stage-2.
\end{itemize}

\textbf{Stage-3.}~In stage-3, in addition to the constructed ImageNet-Instruct-130K, we created a larger image-generated instruction-following dataset, ImageNet-Instruct-1270K, which, in comparison to ImageNet-Instruct-130K, boasts a more diverse array of Prompts and Answers templates (up to 400). The construction of Prompts and Answers involves the direct concatenation of templates with captions.

\section{Experiments}
\label{sec:experiments}

\begin{table*}[htbp]
  \centering
    \begin{tabular}{c|cc|cccccc}
    \toprule
    \multirow{2}[2]{*}{\textbf{Methods}} & \multirow{2}[2]{*}{\textbf{LLM}} & \multicolumn{1}{c|}{\multirow{2}[2]{*}{\textbf{Gen.}}} & \multirow{2}[2]{*}{\textbf{GQA}} & \multirow{2}[2]{*}{\textbf{TextVQA}} & \multirow{2}[2]{*}{\textbf{VQAv2}} & \multirow{2}[2]{*}{\textbf{SciQA-img}} & \multirow{2}[2]{*}{\textbf{OKVQA}} & \multirow{2}[2]{*}{\textbf{VizWizQA}} \\
          &       &       &       &       &       &       &       &  \\
    \midrule
    MiniGPT-4 (7B) & Vicuna-7B & $\times$      & 43.5  & -     & 0.6   & 39.6  & -     & - \\
    InstructBLIP (8B) & Vicuna-7B & $\times$      & 49.2  & -     & -     & 60.5  & -     & 34.5 \\
    Qwen-VL (7B) & Qwen-7B & $\times$      & 59.3  & 50.1  & -     & 67.1  & -     & 35.2 \\
    Qwen-VL-Chat (7B) & Qwen-7B & $\times$      & 57.5  & \textcolor{gray}{61.5} & 78.2  & 68.2  & 56.6  & 38.9 \\
    mPLUG-Owl2 (7B)  & LLaMA-2-7B  & $\times$      & 56.1  & 53.3  & -     & 68.7  & \textbf{57.7} & 54.5 \\
    LLaVA-1.5-hf (7B) & Vicuna-7B & $\times$      & 61.2  & 48.8  & 76.49 & 67.9  & 53.2  & 53.53 \\
    \midrule
\rowcolor{aliceblue}    \textbf{VARGPT(7B+2B)} & \textbf{Vicuna-7B} & $\checkmark$ & \textbf{62.3} & \textbf{54.1} & \textbf{78.4} & \textbf{80.1} & 55.8  & \textbf{56.83} \\
    \bottomrule
    \end{tabular}%
  \caption{ Performance comparison on visual question answering tasks.  We \textcolor{gray}{gray} out the model has trained on the dataset. \textbf{Gen} represents whether the method supports image generation capability.}
  \label{tab:main-2}%
\end{table*}%

\begin{table}[htbp]
  \centering
    \begin{tabular}{rl|cc}
    \toprule
          &       & \textbf{FID}$\downarrow$ & \textbf{CLIP}$\uparrow$ \\
    \midrule
    \multicolumn{1}{c}{\multirow{3}[1]{*}{T.}} & w/o Stage 1. and 2. & 15.4  & 21.3 \\
          & w/o Stage 3. & 20.1  & 20.2 \\
          & w/o Stage 1. & 14.6  & 26.5 \\
          \midrule
    \multicolumn{1}{c}{\multirow{2}[1]{*}{P.}} & Freeze Projector (Stage 3) & 16.2  & 24.5 \\
          & Freeze Visual Decoder (Stage 3) & 17.8  & 22.4 \\
    \midrule
        \rowcolor{aliceblue}  & \textbf{Stage1 +2 +3 (Full)} & \textbf{12.6} & \textbf{27.4} \\
    \bottomrule
    \end{tabular}%
  \caption{Ablation experiments of VARGPT for visual generation. T. denotes `Training' representing ablation studies conducted on the training strategies and P. signifies `Parameters' indicating ablation experiments performed on the parameters fine-tuned during training.}
  \label{tab:ablation-1}%
\end{table}%

\begin{figure*}[t]
  \centering
   \includegraphics[width=1\linewidth]{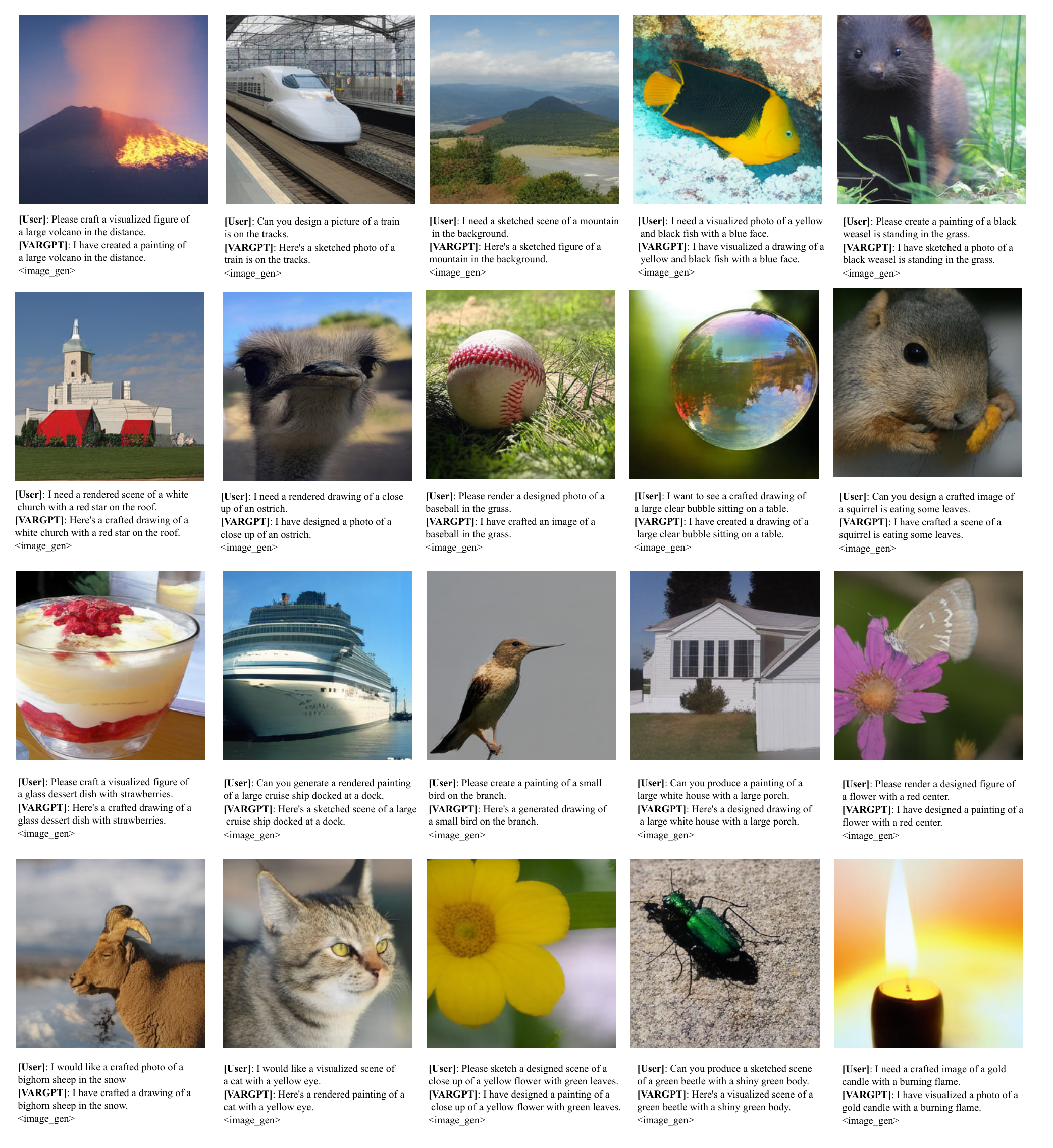}

   \caption{Some generated 256$\times$256 samples by our VARGPT trained on ImageNet-1K. VARGPT supports user text command input and outputs both text and image modal data simultaneously.}
   \label{fig:generate-vis}
\end{figure*}

\begin{table}[htbp]
  \centering

    \begin{tabular}{rl|cc}
    \toprule
          &       & \textbf{MMMU}$\uparrow$ & \textbf{MME}$\uparrow$ \\
    \midrule
    \multicolumn{1}{c}{\multirow{2}[1]{*}{P.}} & Freeze Projector (Stage 2) & 35.19 & 1452.0  \\
          & Freeze LLM in Stage 2 & 33.51 & 1392.5 \\
          \midrule
    \multicolumn{1}{c}{\multirow{3}[1]{*}{D.}} & - LLaVA-OneVision-502K & 33.56 & 1403.5 \\
          & - LLaVA-1.5-665K & 32.62 & 1256.2 \\
          & + ImageNet-Instruct-130K & 36.35 & 1491.2 \\
    \midrule
        \rowcolor{aliceblue}   & \textbf{Full} & \textbf{36.44} & \textbf{1488.8} \\
    \bottomrule
    \end{tabular}%
      \caption{Ablation experiments of VARGPT for visual understanding. P. denotes `Parameters' indicating ablation experiments conducted on the parameters fine-tuned during model training. D. represents `Dataset' signifying ablation studies performed to evaluate data efficiency.
}
  \label{tab:ablation-2}%
\end{table}%

\begin{table}
  \begin{minipage}{0.99\linewidth}
\centering
\scalebox{0.80}{
\begin{tabular}{l p{7.5cm} }
\toprule
 \multicolumn{2}{l}{\bf Visual input example:}  \\
\midrule
&  \includegraphics[height=3cm]{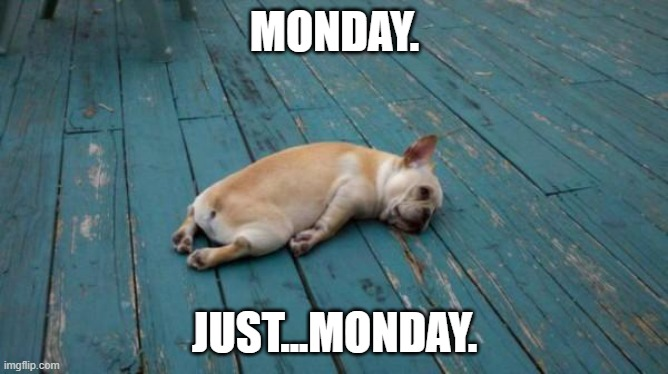} \\

User & Please explain the meme in detail. \\
\midrule
LLaVA-v1.5-7B & The image features a small brown dog lying on the floor, possibly on a wooden surface, with its head up and eyes open. The dog appears to be sleeping or resting, and it is positioned in a way that it seems to be looking up at the camera. The scene is set against a blue background, which adds a contrasting color to the image. The image is accompanied by a caption that reads "Just Monday...", suggesting that the dog is enjoying a lazy day at home. The caption is placed at the bottom of the image, with the dog taking up most of the frame. \\
\midrule
VARGPT & The image depicts a small dog lying on a wooden deck. The dog appears to be resting or sleeping, with its body stretched out on the wooden planks. The dog has a relaxed posture, with its head resting on the wooden surface and its legs stretched out in front of it. The deck is made of wooden planks, and the planks are slightly worn, showing signs of weathering and age. The background of the image is blurred, focusing the viewer's attention on the dog and the wooden deck. Overlaying the image is a text in a speech bubble that reads: "Monday. Just... Monday." The text is in a bold, white font and is positioned in the upper part of the image, above the dog. The speech bubble is floating above the dog, indicating that the text is not part of the actual image but is superimposed digitally. The text "Monday. Just... Monday." is a play on words, combining the common phrase "Just... Monday" with the word "Monday." \textbf{This phrase is often used humorously to express the feeling of being tired or overwhelmed on the first day of the workweek.}\\
\bottomrule
\end{tabular}
}
\captionof{table}{VARGPT demonstrates the capability to comprehend and interpret humorous elements within visual content.}
\label{tab:tricky_example}  
  \end{minipage}
\end{table}

\textbf{Details.}~
For images utilized in image generation tasks, we uniformly resize and crop them to 256x256 pixels, subsequently applying preprocessing techniques consistent with those employed in the VAR~\cite{VAR}.
Regarding images for visual comprehension tasks, we adhere to the preprocessing protocol established in the LLaVA-1.5 framework.
Our language model, visual encoder, and visual feature mapper are initialized using the LLaVA-1.5-7B-hf~\footnote{\url{huggingface.co/llava-hf/llava-1.5-7b-hf}} architecture. The visual decoder is initialized with VAR-d30 parameters, encompassing approximately 2 billion model parameters. The visual generation feature mapper in VARGPT undergoes random initialization and is preliminarily updated during the first stage of pre-training.
We adopt a multi-scale VQVAE~\cite{esser2021tamingtransformershighresolutionimage} similar to VAR~\cite{VAR} for image tokenization, facilitating the next-scale prediction paradigm. A comprehensive summary of the training details across the three stages of our model is presented in Table~\ref{tab:setting}.
During image generation, the top-k and top-p sampling parameters of our VARGPT model are set to 900 and 0.95, respectively. Additionally, the CFG scale parameter is configured at 1.5.

\textbf{Benchmarks.}~
Following common settings~\cite{liu2024llavanext, zhu2023minigpt, liu2024improvedbaselinesvisualinstruction}, 
we evaluate the effectiveness of our VARGPT in visual understanding on a collection of both academic task-oriented benchmarks and recent benchmarks specifically proposed for instruction-following MLLMs, totaling 11 benchmarks:

(1) five multi-modal benchmarks for instruction-following MLLMs, including MMbench-dev (en)~\cite{liu2024mmbenchmultimodalmodelallaround}, SEED-bench~\cite{li2023seedbenchbenchmarkingmultimodalllms}, MMMU~\cite{yue2024mmmumassivemultidisciplinemultimodal}, POPE~\cite{li-etal-2023-evaluating} and MME~\cite{fu2024mmecomprehensiveevaluationbenchmark} benchmarks. For the POPE benchmark, we evaluate on \textit{random}, \textit{popular} and \textit{adversarial} settings with accuracy as metrics;
(2) six visual-centric question-answer benchmarks, including GQA~\cite{hudson2019gqanewdatasetrealworld}, TextVQA~\cite{singh2019vqamodelsread}, VQAv2~\cite{goyal2017makingvvqamatter}, SciQA-img~\cite{lu2022learnexplainmultimodalreasoning}, OKVQA~\cite{marino2019okvqavisualquestionanswering} and VizWizQA~\cite{gurari2018vizwizgrandchallengeanswering}. 
For visual understanding benchmarks, we use the settings in lmms-eval~\cite{zhang2024lmmsevalrealitycheckevaluation} to achieve a unified evaluation.

For visual generation assessment, we constructed an evaluation dataset comprising 50,000 text instructions to assess the model's generative capabilities. We employed CLIP-score to evaluate the CLIP scores between the text instructions and the generated images. Additionally, we utilized the Fréchet Inception Distance (FID) metric to assess the quality of image samples generated by our VARGPT model, which was trained on the ImageNet-1K dataset.

\textbf{Baselines.}~
We conducted a comparative analysis of our VARGPT model against other multimodal large language models of similar scale designed for visual understanding. The comparison included prominent models such as LLaVA-1.5~\cite{Liu2023VisualIT}, MiniGPT-4~\cite{Chen2023MiniGPTv2LL}, and mPLUG-Owl2~\cite{Ye2023mPLUGOwl2RM}, as well as InstructBLIP~\cite{dai2023instructblipgeneralpurposevisionlanguagemodels} and Qwen-VL~\cite{bai2023qwen}.
Furthermore, we extended our comparative study to encompass unified models, including Chameleon~\cite{chameleonteam2024chameleonmixedmodalearlyfusionfoundation}, SEEDLLaMA~\cite{ge2023makingllamadrawseed}, Show-o~\cite{xie2024showo}, and VILA-U~\cite{xie2024showo}. This comprehensive comparison allowed us to evaluate the performance of VARGPT in relation to a diverse range of state-of-the-art models in the field.

\subsection{Main Results}

\noindent\textbf{Evaluation on Multi-modal Benchmarks.}~
We conduct a zero-shot multimodal evaluation and compare our VARGPT with various multimodal models designed for visual understanding. The results are shown in Table~\ref{tab:main-1}.
Based on the results, we have several detailed observations:
(1) It can be observed that our method significantly outperforms most existing MLLMs baselines for visual understanding, including LLaVA-1.5~\cite{Liu2023VisualIT}, MiniGPT-4~\cite{Chen2023MiniGPTv2LL}, InstructBLIP~\cite{dai2023instructblipgeneralpurposevisionlanguagemodels} and Qwen-VL~\cite{bai2023qwen}. 
Our VARGPT achieves the higher performance has been achieved on all benchmarks and some visual hallucination evaluation benchmarks (i.e. POPE), which demonstrates the superiority and generalizability of our method in visual generation.
(2) Although our core architecture for visual understanding is similar to LLaVA-1.5, our method achieves significantly better performance and also supports visual generation in the single large model.
(3) Compared with other unified models (e.g., SEEDLLaMA~\cite{ge2023makingllamadrawseed} and VILA-U~\cite{xie2024showo}) that support generation and understanding, our model naturally supports mixed mode output (while continuously outputting text and images in a conversation), and achieves significantly better visual understanding.
Furthermore, we conducted sample analyses on the LLaVA-Bench benchmark, with partial results presented in Table~\ref{tab:tricky_example}. Observations indicate that, compared to LLaVA-1.5, our approach demonstrates superior image comprehension capabilities and exhibits an enhanced ability to discern and analyze humorous elements within images.

\noindent\textbf{Evaluation on visual question answering tasks.}~
We compare various visual question answering tasks with existing methods, with the results shown in Table~\ref{tab:main-2}.
As shown in Table~\ref{tab:main-2}, we have several observations: 
(1) Our VARGPT consistently achieves optimal results on most understanding benchmarks, surpassing MLLMs designed for visual understanding with the same parameter size.
This further demonstrates the effectiveness of our VARGPT;
(2) In addition to achieving significant comprehension abilities (such as 12.2\% higher than LLaVA-1.5 on the SciQA-img benchmark), VARGPT can also support visual generation capabilities compared to these baselines.

\noindent\textbf{Evaluation on Instruction-to-image Task.}~
To evaluate the visual generation capabilities of our VARGPT, we construct an instruction-based question-answering generation evaluation dataset comprising 50,000 samples. The instructional descriptions in this dataset are derived from ImageNet-1K image captions, with a constraint of 50 samples per category to ensure balanced representation across classes.
To quantitatively assess the instruction-following ability of VARGPT, we evaluated two key metrics: (1) the FID score between the 50,000 generated images and the ImageNet-1k dataset, and (2) the CLIP score between the instructions and the generated images by CLIP model. The results of this evaluation are presented in Table \ref{tab:ablation-1}.
Furthermore, we provide the visualization of the images and dialogues generated by VARGPT in Figure \ref{fig:generate-vis}. Observational analysis reveals that our VARGPT is capable of producing high-quality images that closely adhere to the given instructions.
Notably, VARGPT demonstrates the ability to seamlessly integrate text description and image generation within a single dialogue, utilizing a single unified model for both multimodal input and output. This capability further underscores VARGPT's unique advantage in unifying visual generation and comprehension tasks.
The image generation dataset utilized in VARGPT (1.28M ImageNet) is substantially smaller and of lower quality compared to those employed by other unified models (e.g., Show-1: 36M, VILA-U: 15M, Liquid: 30M images). Consequently, VARGPT's image generation performance currently lags behind these approaches. However, the potential for quality improvement through data scaling presents a promising avenue for future research and development.

\begin{figure}
    \centering
    \begin{subfigure}{0.49\linewidth}
      \includegraphics[width=1\linewidth]{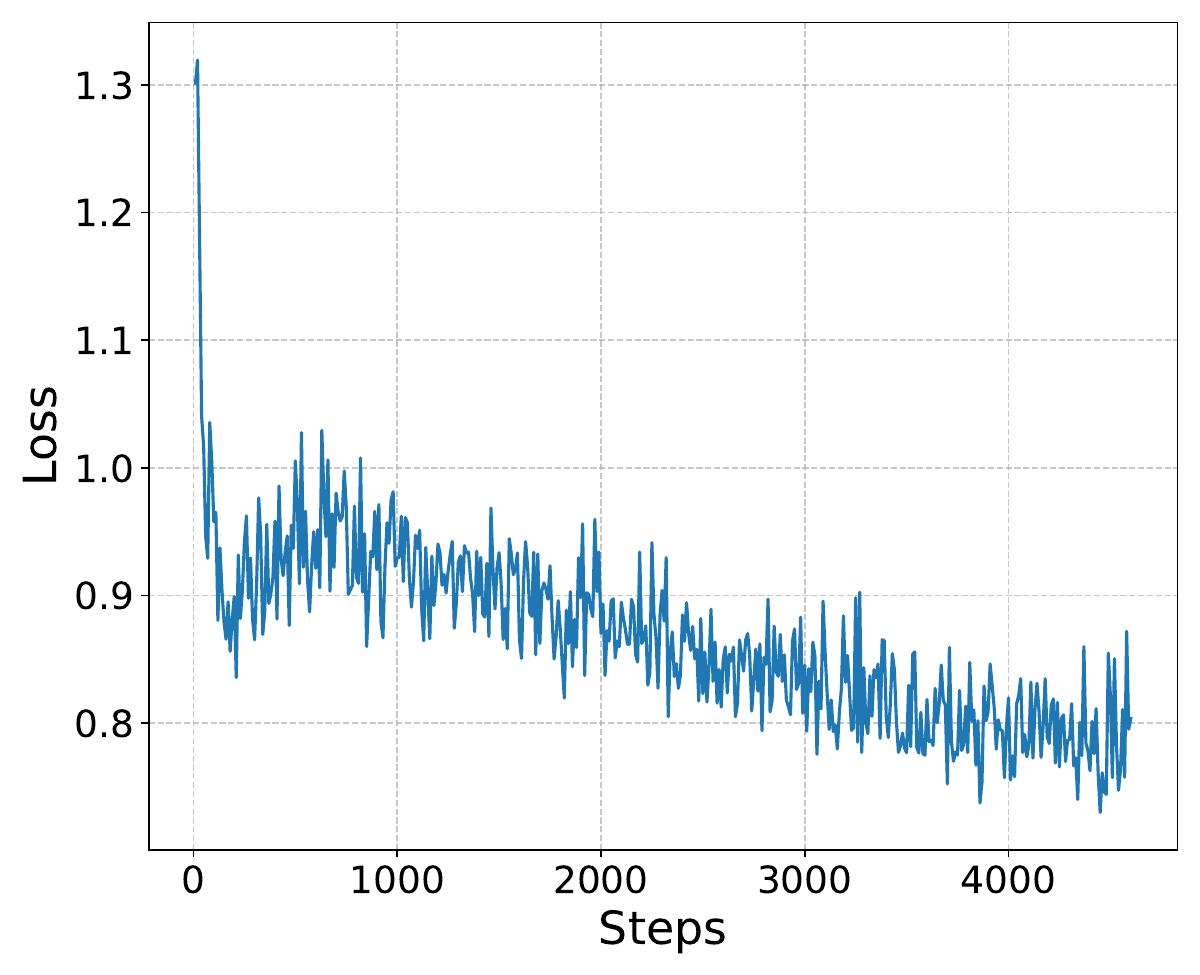}
      \caption{Stage-2 training loss curve.}
      \label{fig:exp-1}
    \end{subfigure}
    \hfill
    \begin{subfigure}{0.49\linewidth}
      \includegraphics[width=1\linewidth]{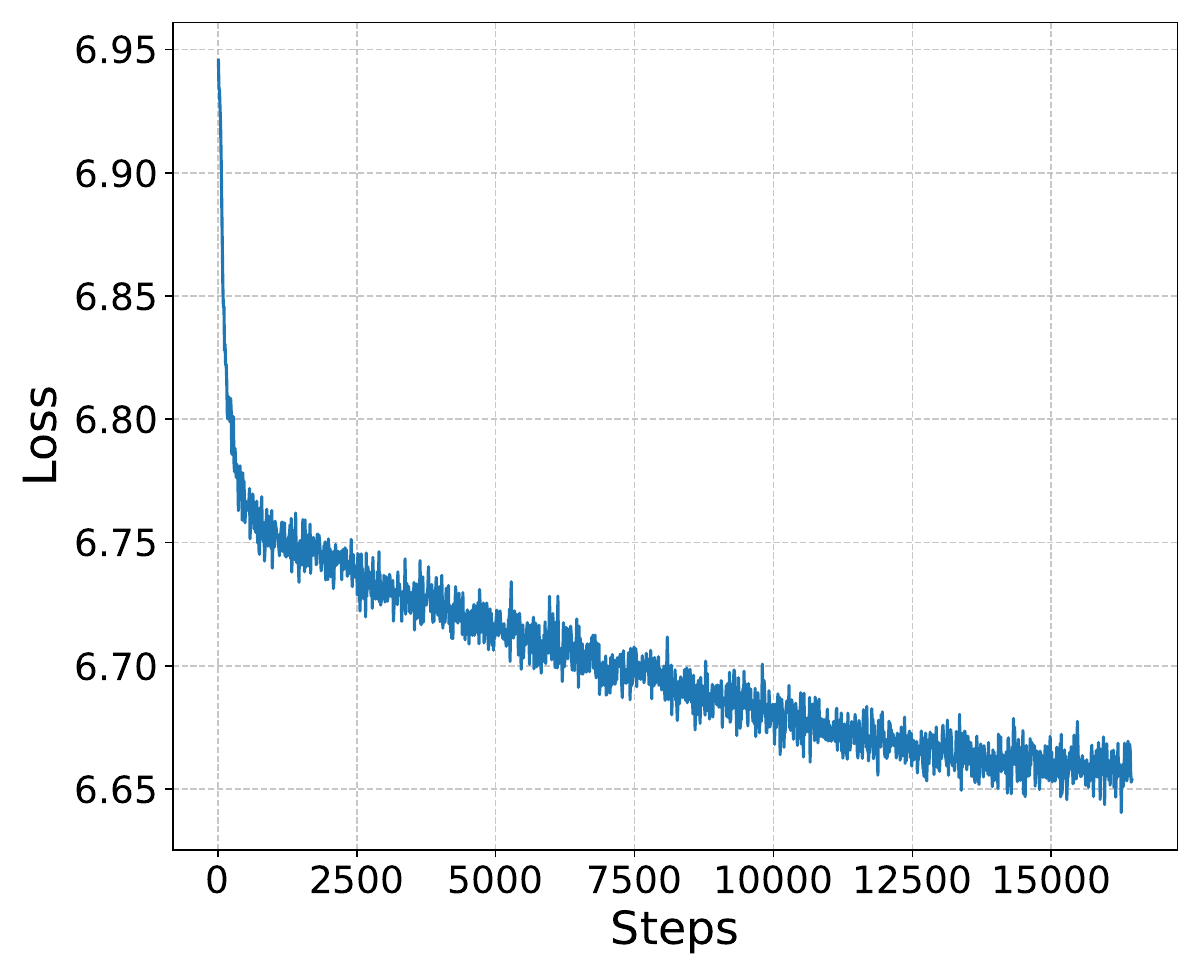}
      \caption{Stage-3 training loss curve.}
      \label{fig:exp-2}
    \end{subfigure}
    \caption{Loss curves of training stages 2 and 3 for VARGPT.}
        \label{fig:exp-loss-curve}
        
  \end{figure}

\subsection{Method Analysis}
We conduct ablation experiments on our VARGPT in terms of model parameters, training settings, and data efficiency to evaluate the effectiveness of various components in detail. Specifically, we evaluate the effectiveness of the components by removing the specific settings, as shown in Table~\ref{tab:ablation-1} and~\ref{tab:ablation-2}.

\noindent\textbf{Effect of the Training Strategies on Generation.}~
As shown in Table~\ref{tab:ablation-1}, 
the omission of any single stage or combination of stages in our training protocol results in a significant deterioration of our model's visual generation performance. Notably, the removal of the third stage, which involves instruction fine-tuning, leads to a substantial decline in both the quality of generated images and the model's ability to adhere to given instructions.
These findings underscore the crucial role that each of the three training stages plays in enhancing the model's visual generation quality and text-to-image capabilities. Moreover, we conducted additional experiments wherein we selectively froze the mapper and visual decoder parameters during the third stage of training. Our observations indicate that the absence of fine-tuning for these components during the third stage also results in performance degradation.
Collectively, these results provide compelling evidence for the efficacy of our three-stage training strategy for VARGPT. The consistent performance decline observed across various ablation scenarios reinforces the importance of each proposed component and stage.

\noindent\textbf{Effect of the Training Strategies on Understanding.}~
To assess the effectiveness of our training strategy on visual comprehension capabilities, we conducted ablation studies by selectively freezing components during the second stage of training. Specifically, we performed separate experiments where we froze either the mapper or the LLM backbone while proceeding with instruction fine-tuning in the second stage.
As illustrated in Table \ref{tab:ablation-2}, we observed a significant performance decline in both scenarios. These results provide further validation of the efficacy of our training strategy in enhancing visual comprehension abilities. 
This empirical evidence underscores the importance of allowing both the mapper and the LLM backbone to adapt during the instruction fine-tuning phase, highlighting the synergistic effect of our proposed training methodology on the model's overall visual understanding capabilities.

\noindent\textbf{Effect of the Data Efficiency on Understanding.}~
Furthermore, we conduct experiments to analyze the mixed dataset used in our second stage of training. The results are presented in Table~\ref{tab:ablation-2}. We can observe that removing either the 502K or 665K comprehension datasets negatively impacts the model's understanding performance. Conversely, when we further incorporated our constructed instruction dataset for generation, it enhanced the model's ability to differentiate between comprehension and generation instructions and accurately improved VARGPT's capability to output special tokens for visual generation (i.e., \texttt{<image\_gen\_start>}, \texttt{<image\_gen>}, and \texttt{<image\_gen\_end>}) without significantly affecting its comprehension performance.

\noindent\textbf{Visualization of Training Loss Curve.}~ 
We further illustrate the loss curves for the second and third stages of our model's training process in Figure~\ref{fig:exp-loss-curve}. The observed trends in these loss curves demonstrate a generally reasonable and consistent decline, providing empirical support for the effectiveness of our learning strategy.
An analysis of these curves reveals that the training loss exhibits a principled decrease over time, which to a considerable extent corroborates the efficacy of our proposed learning approach. 
Moreover, a closer examination of the third-stage loss curve suggests that there remains significant potential for further optimization of the model's visual generation capabilities.
This observation indicates that extending the training duration and expanding the training dataset could yield additional improvements in visual generation performance during the third stage.

\section{Conclusion, Limitation and Future Work}
\label{sec:conclusion}

\textbf{Conclusion.}~
This work introduces VARGPT, a novel MLLMs that successfully integrates visual understanding and generation within a unified autoregressive framework. By employing innovative next-token and next-scale prediction paradigms, VARGPT extends the capabilities of traditional MLLMs to include efficient visual autoregressive generation. The model's three-stage training pipeline, utilizing specially curated datasets, enables effective alignment between visual and textual features, enhancing both understanding and generation capabilities.
VARGPT demonstrates superior performance compared to existing models like LLaVA-1.5 across various vision-centric tasks. Moreover, it exhibits remarkable proficiency in autoregressive visual generation and text-to-image synthesis. These achievements underscore VARGPT's versatility and potential to advance the field of multimodal AI, providing meaningful exploration for future research on unified multimodal models

\textbf{Limitation.}~
\label{sec:limitation}
(1) As our visual generation dataset primarily sources images from ImageNet, there remains a discernible quality gap between VARGPT and certain diffusion models, such as SDv2.1~\cite{rombach2021highresolution} and more advanced models like FLUX~\cite{flux2023}, which are pre-trained on extensive high-quality image datasets. This disparity is primarily attributable to differences in training data. Furthermore, throughout our training process, the resolution for generated images was consistently set at 256×256 pixels. Consequently, the current version of VARGPT exclusively supports autoregressive generation of images at this resolution. 
(2)  While VARGPT demonstrates initial proficiency in instruction understanding and instruction-to-image generation, effectively following user input instructions in the majority of scenarios, there are instances where certain nuanced details within instructions may not be adequately reflected in the generated images. This limitation manifests in some cases, indicating room for improvement in the model's ability to comprehensively capture and render intricate instructional details.

\textbf{Future Work.}~
\label{sec:Future_Work}
(1) To achieve superior image generation quality and support higher resolution outputs, we plan to improve the next-scale prediction model architecture, expand the image dataset, enhance image quality, and implement dynamic resolution capabilities.
(2) In subsequent versions of VARGPT, we intend to explore support for unified autoregressive video understanding and generation. 

\maketitlesupplementary

\section{Details of Model Architecture}
\label{sec:appendix-model-details}
\subsection{Visual Decoder.}~
Our VARGPT incorporates a visual decoder with 2 billion parameters dedicated to visual generation. The visual features generated by the LLM are input into a generative feature mapper before being fed forward into the visual decoder. The two mappers for visual generation and the visual decoder can be conceptualized as translators, converting the features output by the LLM into visual information.

\textbf{In our experiments, we observe that sharing identical parameters between visual generation and text generation led to a phenomenon of knowledge conflict. }When we enhanced the model's comprehension capabilities, its visual generation performance declined, and conversely, improving the model's visual generation abilities resulted in a decrease in its understanding capacity. Consequently, our VARGPT utilizes a 2 billion parameter visual decoder as specialized parameters for implementing visual generation.

This architectural decision allows VARGPT to maintain distinct parameter spaces for visual generation and textual processing, mitigating the observed knowledge conflict and enabling the model to excel in both domains simultaneously. The dedicated visual decoder serves as a crucial component in bridging the gap between linguistic and visual representations, facilitating the model's capacity to generate high-quality visual outputs based on textual inputs. Our visual decoder is composed of 30 Transformer blocks, each incorporating 30 attention heads. The decoder's width is set at 1920, and it employs adaptive normalization (AdaLN).

\subsection{ Multi-scale Tokenizer.}~
We employ a multi-scale VAE~\cite{esser2021tamingtransformershighresolutionimage, VAR} architecture analogous to that used in VAR, which utilizes a vanilla VQVAE~\cite{esser2021tamingtransformershighresolutionimage} structure complemented by a multi-scale quantization scheme incorporating K additional convolutions (amounting to 0.03M additional parameters).
The multi-scale tokenizer features a vocabulary size of 4090 and was trained on the OpenImages~\cite{Kuznetsova_2020} dataset. It operates with a spatial downsampling factor of 16$\times$.

\subsection{ Position Encoding.}~
VARGPT's LLMs employs the original positional encoding from Vicuna~\cite{chiang2023vicuna}. However, when image features are input into the visual decoder, we additionally incorporate absolute positional encoding, which serves to differentiate visual generation tokens at various scales.

\subsection{Design of Projectors.}~
For the two newly added projectors designed to implement visual generation feature mapping, we structured each projector as a single linear layer followed by a GELU activation function, culminating in another linear layer. This architecture allows for efficient and non-linear transformation of features between the language and visual domains.

\begin{figure}[t]
  \centering
   \includegraphics[width=1\linewidth]{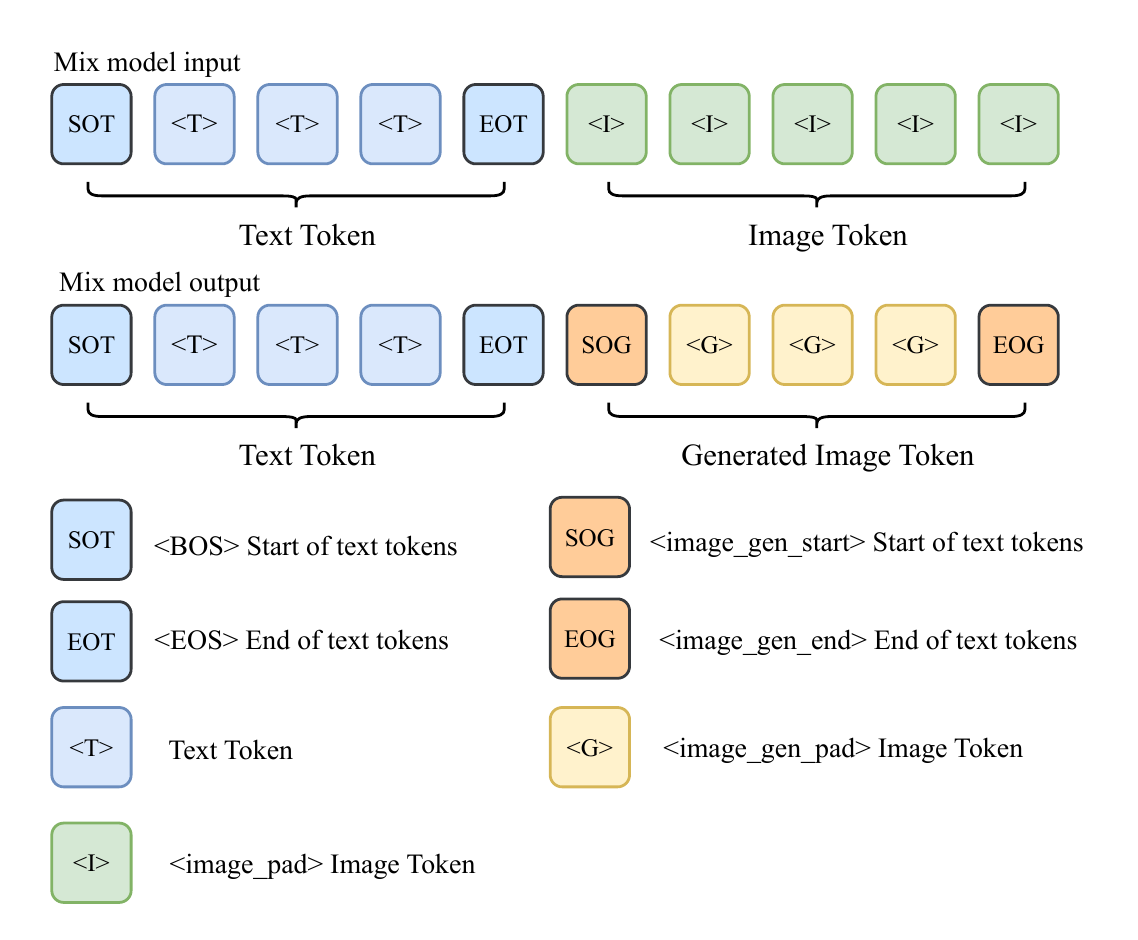}
   \caption{ Illustration of the proposed unified prompting format.} 
   \label{fig:vargpt_appendix_token}
\end{figure}

\subsection{ Special Token and Prompt for Generation.}~
We supplemented with several special tokens to facilitate VARGPT's mixed-modal output capabilities. The distribution of these output tokens during mixed-modal generation is illustrated in Figure~\ref{fig:vargpt_appendix_token}. This visualization provides insight into how the model transitions between textual and visual modalities in its generative process.

\begin{figure*}[htp]
  \centering
   \includegraphics[width=1\linewidth]{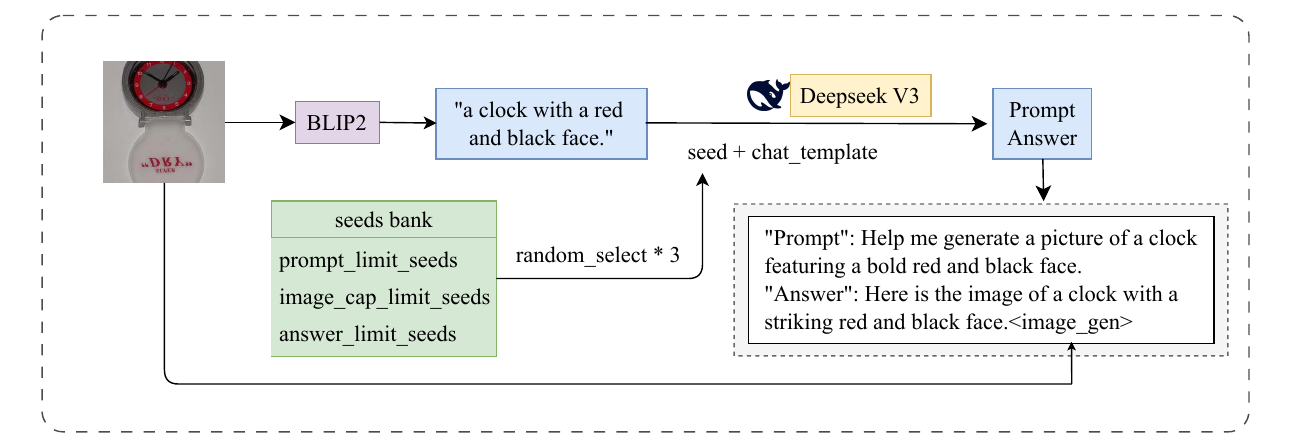}
   \caption{ Illustration of the proposed unified visual generation instruction fine-tuning dataset construction.} 
   \label{fig:vargpt_appendix_datasets}
\end{figure*}

\subsection{ Classifier-free Guidance.}~
Classifier-free guidance (CFG) significantly enhances the capability of generative diffusion models to produce samples of exceptionally high fidelity. This approach integrates conditional generative models with the distribution estimation of unconditional models trained concurrently, thereby improving the overall quality of generation.
Inspired by DALL-E 2~\cite{ramesh2022hierarchicaltextconditionalimagegeneration}, VAR~\cite{VAR} and VAR-CLIP~\cite{zhang2024varcliptexttoimagegeneratorvisual}, we employ Gaussian noise features as conditional input to simulate unconditional generation. Subsequently, we derive the final distribution of image token outputs by subtracting the probability of unconditional generation from the logits distribution of conditional generation. 

Specifically, assuming that the image features obtained through LLMs are denoted as $\mathbf{H}^i$, we derive the translated visual generation features through a mapping layer as $\mathbf{H}^g = \operatorname{Projector}(\mathbf{H}^i)$. Subsequently, we employ the Classifier-Free Guidance (CFG) strategy to obtain the final features of the visual generation tokens, which can be expressed as:

\begin{equation}
        \begin{aligned}
        \mathbf{R}_t & \sim  p_{\{\theta, \phi\}} (\mathbf{R}_t \mid \mathbf{H}^g, \mathbf{R}_{<t});            \\
        \mathbf{R}_e & \sim  p_{\{\theta, \phi\}} (\mathbf{R}_t \mid \mathbf{H}^e, \mathbf{R}_{<t});    \\
        \mathbf{R}_t & = (1+\lambda) \mathbf{R}_t - \lambda \mathbf{R}_e
        \end{aligned}
\end{equation}

where $\mathbf{H}^e$ represents randomly initialized Gaussian noise, serving as the unconditional feature, and $\lambda$ denotes the scale hyperparameter for the CFG strategy.

This formulation allows for the integration of both conditional and unconditional information in the generation process, potentially leading to more controlled and diverse visual outputs.

\section{Details of Dataset Construction}
\label{sec:appendix_data_construction}
\begin{figure*}[htp]
  \centering
   \includegraphics[width=0.9\linewidth]{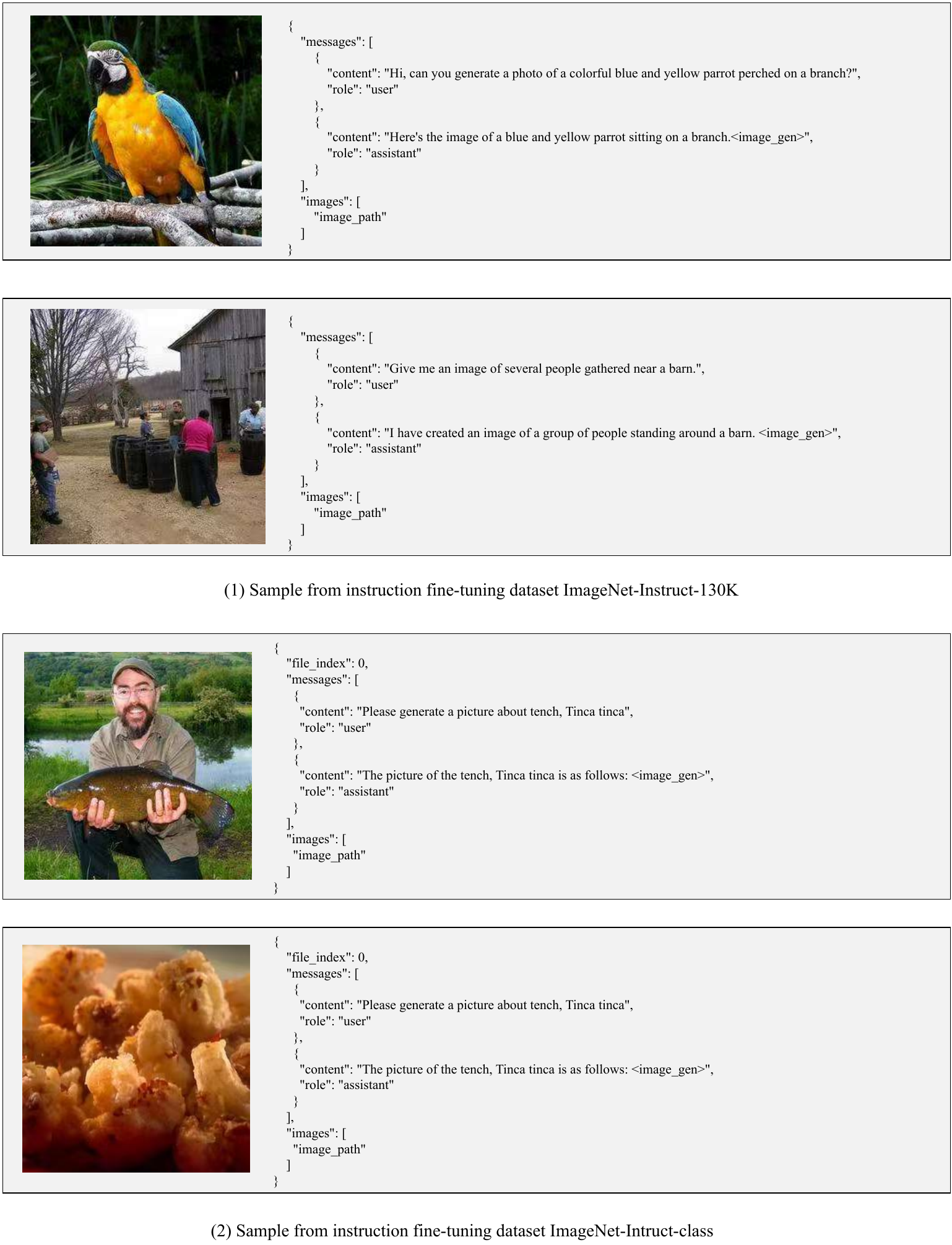}

   \caption{Example of the instruction fine-tuning dataset we collected and constructed.}
   \label{fig:instruction-following-datasets}
\end{figure*}

We first input images from ImageNet~\cite{deng2009imagenet} into the BLIP2~\cite{li2023blip} model to generate captions, forming image-caption pairs, as shown in Figure~\ref{fig:vargpt_appendix_datasets}. Using DeepSeek V3~\cite{deepseekai2024deepseekv3technicalreport}, we generated prompt\_limit\_seeds, image\_cap\_limit\_seeds, and answer\_limit\_seeds to ensure diversity and constraints in the generated prompts and answers. For each image-caption pair, we randomly select different seeds from these sets and populate them into a chat\_template to serve as input for DeepSeek V3. After DeepSeek V3 generates the prompts and answers, we concatenate the answer with a special token <image\_gen> to create a single visual generation instruction fine-tuning dataset sample. By iterating over all image-caption pairs, we obtain the complete dataset.
We show some examples of the visual generation instruction fine-tuning dataset we constructed in Figure~\ref{fig:instruction-following-datasets}.

\section{Qualitative Study for Understanding}

To further evaluate our method's effectiveness in visual understanding, we conducted experiments on LLaVA-Bench~\cite{Liu2023ImprovedBW}, which consists of 24 distinct images with expert-annotated descriptions and corresponding evaluation questions. 
In alignment with previous studies~\cite{Yin2023WoodpeckerHC, Leng2023MitigatingOH, Chen2024HALCOH}, we employed this benchmark for qualitative assessment. 
The visual results are presented in Table~\ref{tab:tricky_example-1}, ~\ref{tab:tricky_example-2}, ~\ref{tab:tricky_example-3},~\ref{tab:tricky_example-4}, and \ref{tab:tricky_example-5}.
It can be observed that despite VARGPT adhering to the core architecture of LLaVA-1.5 for understanding tasks and employing a training strategy similar to LLaVA-1.5, it demonstrates a marked superiority in comprehension capabilities compared to LLaVA-1.5-7B.

\section{Qualitative Study for Generation}
\begin{figure*}[htp]
  \centering
   \includegraphics[width=0.9\linewidth]{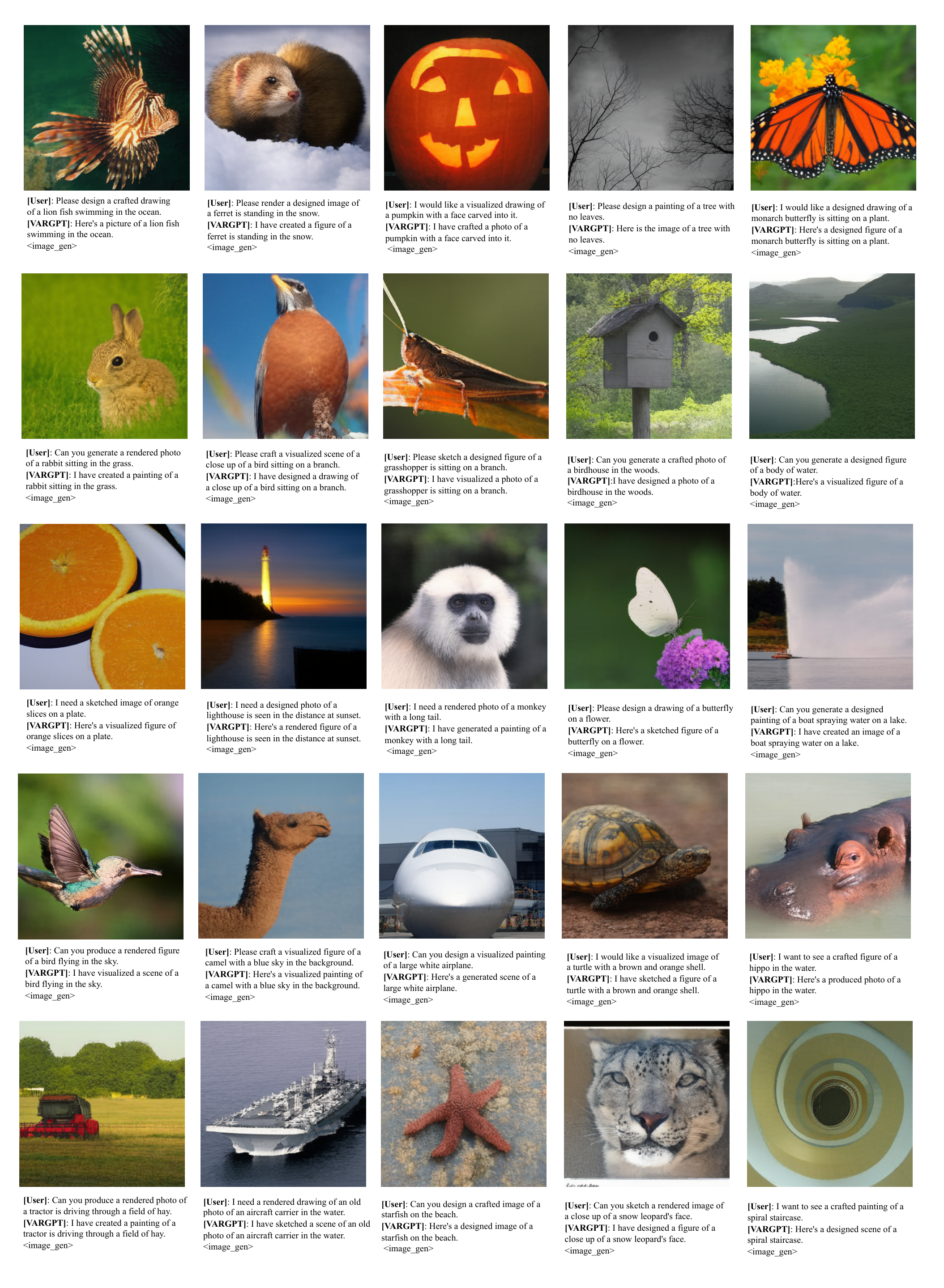}

   \caption{Some generated 256$\times$256 samples by VARGPT trained on ImageNet~\cite{deng2009imagenet}. VARGPT supports user text command input and outputs both text and image modal data simultaneously.}
   \label{fig:instruction-following-generate-1}
\end{figure*}

\begin{figure*}[htp]
  \centering
   \includegraphics[width=0.9\linewidth]{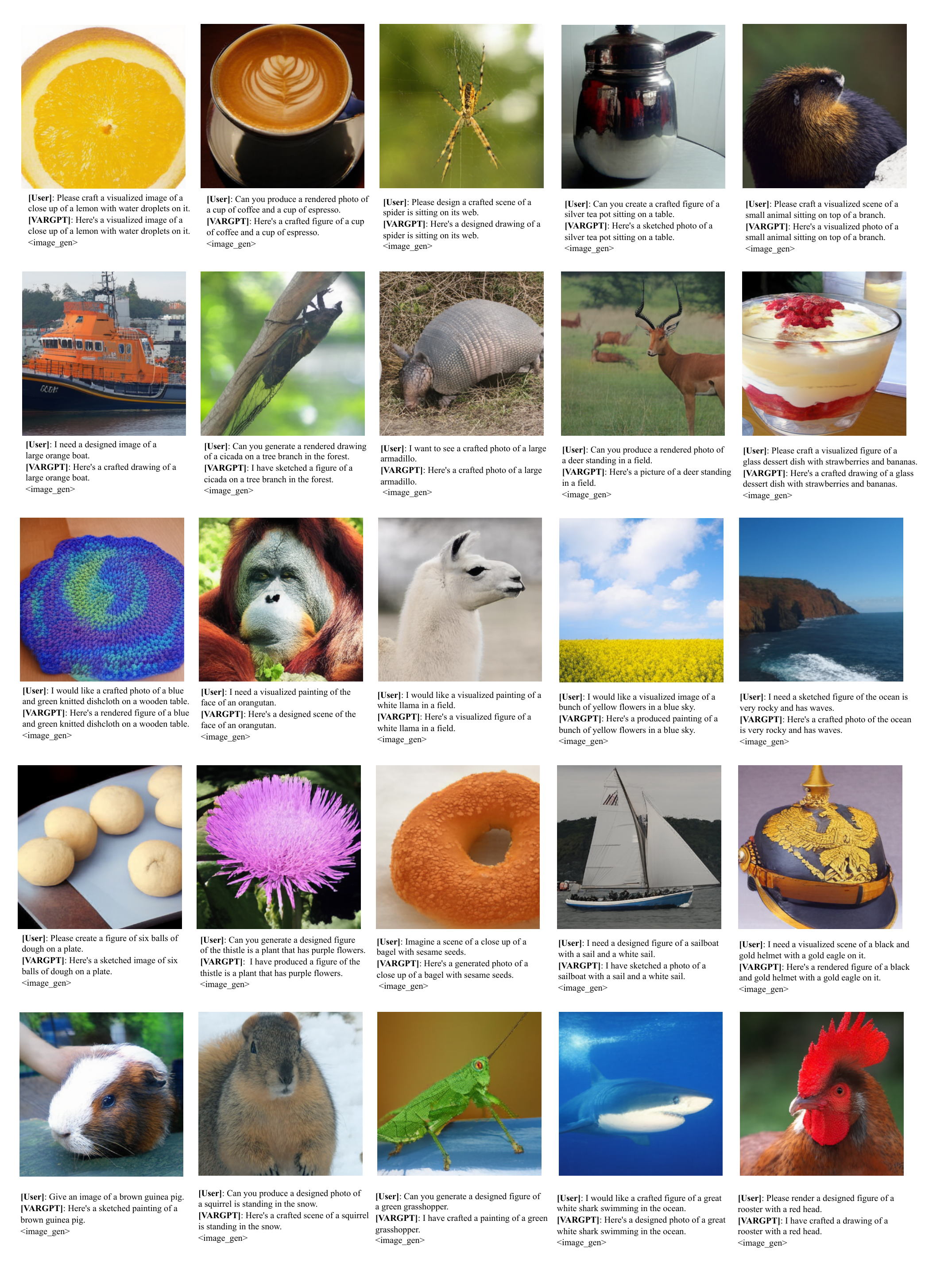}

   \caption{Some generated 256$\times$256 samples by VARGPT trained on ImageNet~\cite{deng2009imagenet}. VARGPT supports user text command input and outputs both text and image modal data simultaneously.}
   \label{fig:instruction-following-generate-2}
\end{figure*}

In Figures~\ref{fig:instruction-following-generate-1} and~\ref{fig:instruction-following-generate-2}, we present additional results demonstrating our VARGPT's capability to output mixed-modality data from various datasets, as well as a quantitative assessment of the quality of instruction-generated images. We observe that our approach naturally supports mixed-modal data, seamlessly embedding visual generation tokens within text-generated tokens, while also producing high-quality image content. These findings underscore the significant effectiveness of our VARGPT in unifying understanding and generation tasks.
The ability to integrate visual and textual elements coherently within a single model output demonstrates the versatility of VARGPT. The high quality of the generated images, as evidenced by quantitative evaluations, further reinforces the model's proficiency in visual synthesis. This seamless integration of multimodal capabilities within a unified framework, offering potential applications across various domains that require both language understanding and visual generation.

\begin{table}
  \begin{minipage}{0.99\linewidth}
\centering
\scalebox{0.80}{
\begin{tabular}{l p{7.5cm} }
\toprule
 \multicolumn{2}{l}{\bf Visual input example:}  \\
\midrule
&  \includegraphics[height=4.5cm]{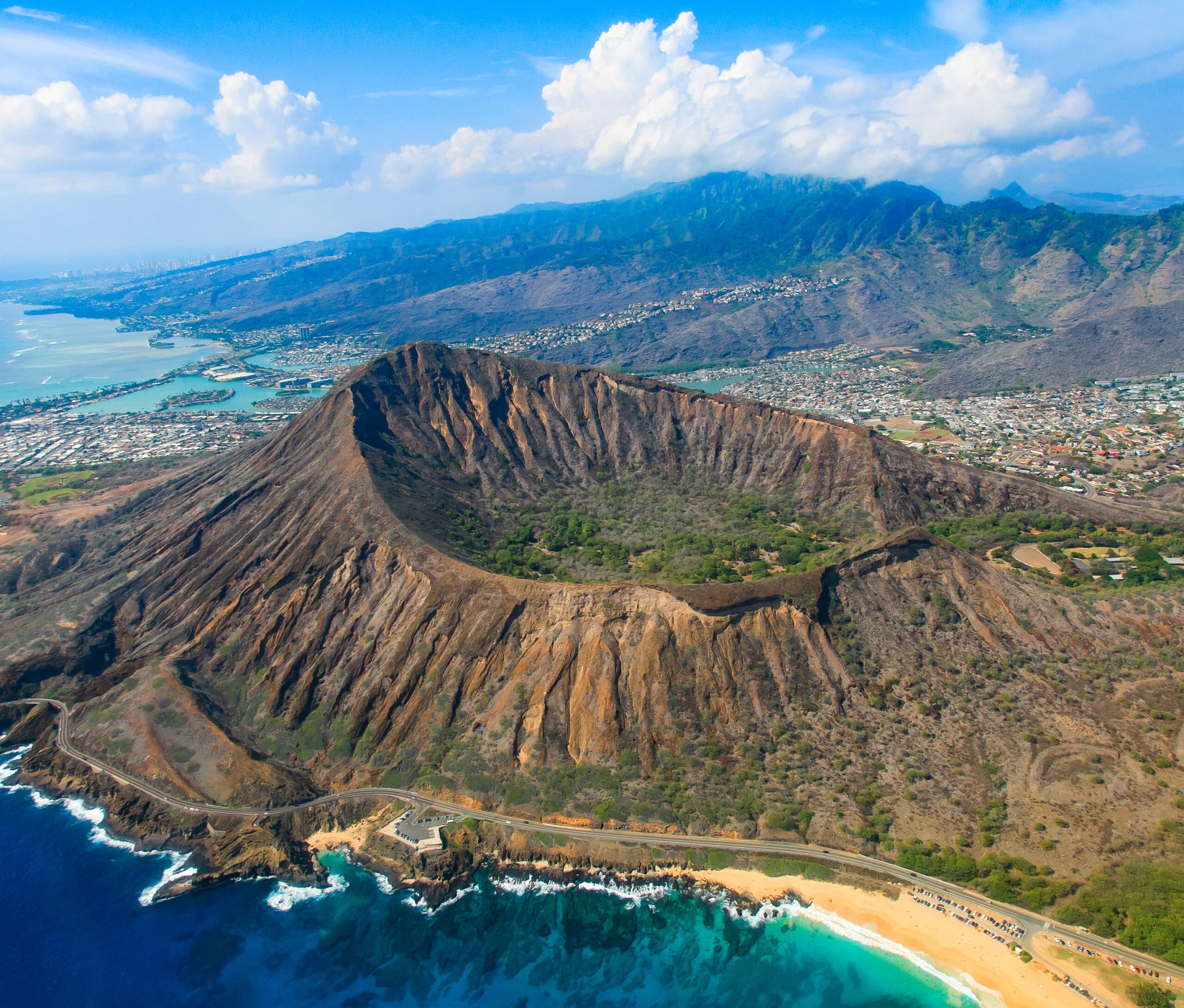} \\

User & What are the possible reasons of the formation of this sight? \\
\midrule
LLaVA-v1.5-7B & The formation of the sight, which is a large, rocky mountain with a lush green hillside and a beach, can be attributed to several factors. Firstly, the mountain's formation is likely due to tectonic forces, such as the collision or separation of tectonic plates, which have resulted in the formation of the mountain range. Secondly, the lush green hillside and the beach are the result of the local climate and geological processes. The presence of a beach suggests that the area is near a coastline, and the lush green hillside could be due to the presence of a water source, such as a river or a waterfall, which has created a fertile environment for plant growth. Additionally, the mountain's unique shape and the presence of a road on its side could be the result of human intervention, such as the construction of roads and infrastructure to facilitate transportation and access to the area. \\
\midrule
VARGPT & The sight of a large mountain with a road running through it and a beach nearby is a result of geological processes that have shaped the landscape over time. Some possible reasons for the formation of this sight include:1. Tectonic forces: The movement of the Earth's crust can lead to the formation of mountains, valleys, and other geological features. The mountain could have been formed through tectonic forces such as the collision of two tectonic plates or the uplift of the Earth's crust.2. Weathering and erosion: The mountain could have been shaped by the forces of weathering and erosion, which break down the rock and soil, leaving behind a rugged landscape. The road running through the mountain could be a result of human intervention to connect different areas or to facilitate transportation.3. Coastal erosion: The beach and the surrounding area could have been shaped by the forces of coastal erosion, which is the process of the sea wearing away the land. The beach could have been formed by the deposition of sediments from the sea, while the surrounding area could have been shaped by the erosion of rock and soil.4. Climate and weather patterns: The vegetation on the mountain and the surrounding area could be influenced by the local climate and weather patterns, which can affect the growth and distribution of plants and animals in the region.\\
\bottomrule
\end{tabular}
}
\captionof{table}{The case of visual understanding in VARGPT shows that our VARGPT has achieved better understanding performance.}
\label{tab:tricky_example-1}  
  \end{minipage}
\end{table}

\begin{table}
  \begin{minipage}{0.99\linewidth}
\centering
\scalebox{0.80}{
\begin{tabular}{l p{7.5cm} }
\toprule
 \multicolumn{2}{l}{\bf Visual input example:}  \\
\midrule
&  \includegraphics[height=6cm]{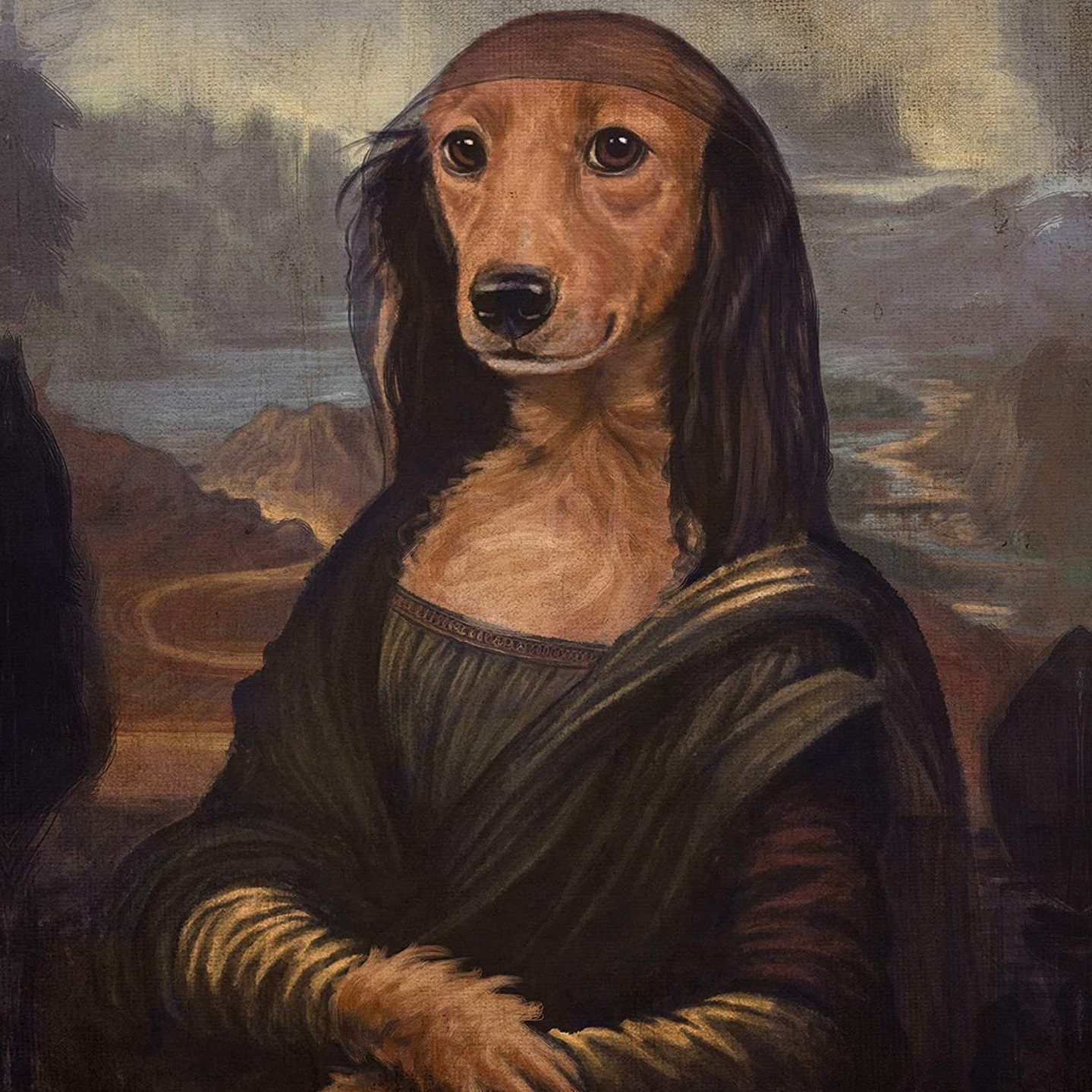} \\

User & Discuss how this creative twist on a classic work of art might be interpreted differently by various audiences. \\
\midrule
LLaVA-v1.5-7B & The creative twist on a classic work of art, featuring a dog wearing a Renaissance-style dress and a beret, can be interpreted differently by various audiences. Some might find it amusing and lighthearted, as it combines the seriousness of a historical painting with the playfulness of a dog dressed up in human attire. Others might view it as a commentary on the human-animal relationship, suggesting that dogs can be just as expressive and creative as humans. Additionally, some might perceive it as a critique of the traditional art world, as it challenges the norms of what is considered \"art\" by incorporating a non-human subject and a more contemporary, humorous element. Overall, the interpretation of this creative twist on a classic work of art can vary greatly depending on the viewer's perspective and personal experiences. \\
\midrule
VARGPT & This creative twist on a classic work of art, where a dog is dressed in a Mona Lisa-style outfit and painted in a similar style, could be interpreted in various ways by different audiences. Some might find it amusing and lighthearted, as it plays on the juxtaposition of a serious artwork with a playful, everyday subject. Others might appreciate the skillful use of artistic techniques to transform a simple image of a dog into a piece of art that can be appreciated on multiple levels. Some might even see it as a commentary on the universal appeal of art, which can be appreciated by both humans and animals. The interpretation would depend on the viewer's personal experiences, artistic background, and sense of humor.\\
\bottomrule
\end{tabular}
}
\captionof{table}{The case of visual understanding in VARGPT shows that our VARGPT has achieved better understanding performance.}
\label{tab:tricky_example-2}  
  \end{minipage}
\end{table}

\begin{table}
  \begin{minipage}{0.99\linewidth}
\centering
\scalebox{0.80}{
\begin{tabular}{l p{7.5cm} }
\toprule
 \multicolumn{2}{l}{\bf Visual input example:}  \\
\midrule
&  \includegraphics[height=3.5cm]{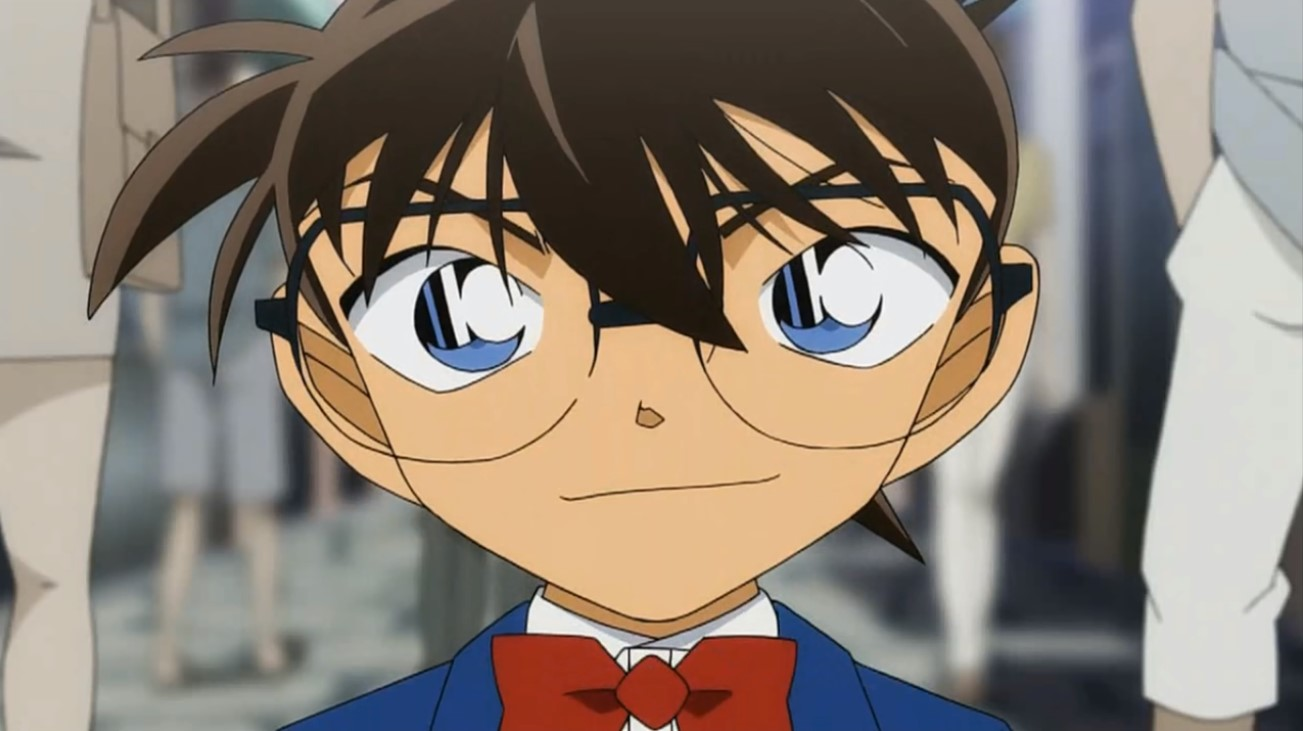} \\

User & What's the personality of this character? Explain what elements or aspects of the character's design may have contributed to its popularity. \\
\midrule
LLaVA-v1.5-7B & The character in the image is a young boy with a distinctive appearance, wearing glasses, a bow tie, and a suit. This unique combination of elements, along with his smiling expression, suggests that the character is friendly, approachable, and confident. The character's design, which includes the bow tie and glasses, adds a touch of sophistication and charm to his overall look. These elements may have contributed to the character's popularity, as they make him stand out and create a memorable impression. The character's personality and design have likely resonated with audiences, making him a beloved and recognizable figure in popular culture. \\
\midrule
VARGPT & The character's personality seems to be confident, intelligent, and perhaps a bit mischievous, as suggested by the slight smirk and the way the eyes are directed slightly upwards and to the side, giving an impression of a playful or slightly rebellious attitude. The glasses and bow tie suggest a formal or professional setting, which could imply that the character is a well-educated individual. The design elements that may have contributed to its popularity include the expressive eyes, the confident stance, and the stylish yet formal attire.\\
\bottomrule
\end{tabular}
}
\captionof{table}{The case of visual understanding in VARGPT shows that our VARGPT has achieved better understanding performance.}
\label{tab:tricky_example-3}  
  \end{minipage}
\end{table}

\begin{table}
  \begin{minipage}{0.99\linewidth}
\centering
\scalebox{0.80}{
\begin{tabular}{l p{7.5cm} }
\toprule
 \multicolumn{2}{l}{\bf Visual input example:}  \\
\midrule
&  \includegraphics[height=4cm]{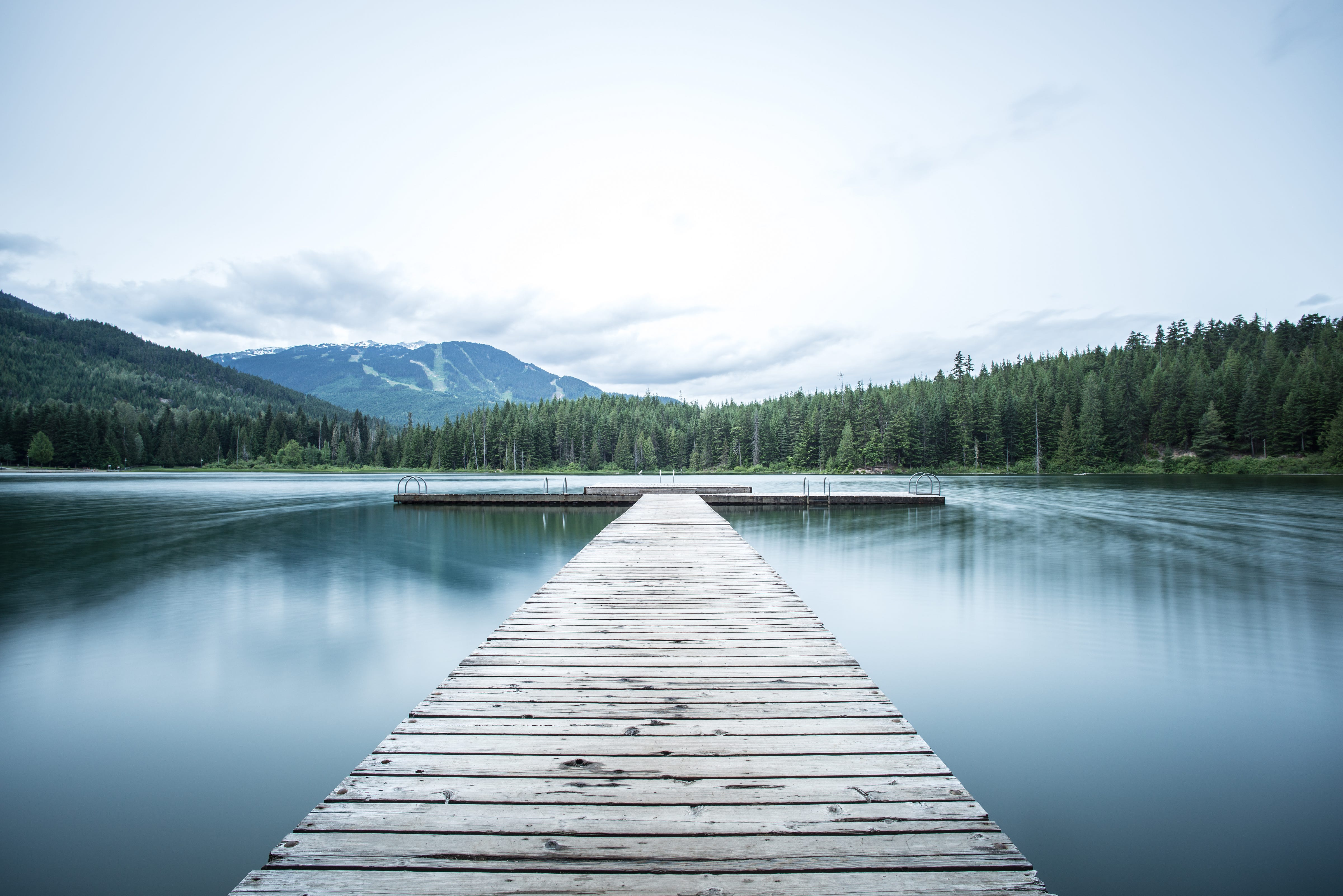} \\

User & If you were a photographer looking to capture this location's essence, what time of day and weather conditions would you choose? Describe the reasons behind your choice. \\
\midrule
LLaVA-v1.5-7B & To capture the essence of this location, which features a pier extending over a lake with mountains in the background, I would choose to photograph during the golden hour, which is the time just after sunrise or before sunset. This time of day provides soft, warm light that enhances the natural beauty of the scene, creating a serene and picturesque atmosphere. Additionally, the golden hour often results in softer shadows and more dramatic lighting, which can accentuate the pier's structure and the surrounding landscape.As for the weather conditions, I would prefer a clear day with a gentle breeze, as it would provide a pleasant and comfortable environment for both the photographer and the subjects. A clear day with good visibility would allow for better composition and capture the stunning mountain backdrop in its full glory. A gentle breeze would also add a touch of dynamism to the scene, making it more visually engaging and interesting. \\
\midrule
VARGPT & I would choose to capture this location's essence during the early morning, specifically during the golden hour, which is the period shortly after sunrise or before sunset. This time of day is known for its soft, warm light and long shadows, which can enhance the tranquility and beauty of the scene. The weather conditions would be clear and calm, with a gentle breeze to create a sense of movement in the water and to soften the reflections on the water's surface. The overcast sky would add a sense of depth and contrast to the scene, while the mist or fog would create a serene and ethereal atmosphere.\\
\bottomrule
\end{tabular}
}
\captionof{table}{The case of visual understanding in VARGPT shows that our VARGPT has achieved better understanding performance.}
\label{tab:tricky_example-4}  
  \end{minipage}
\end{table}

\begin{table}
  \begin{minipage}{0.99\linewidth}
\centering
\scalebox{0.80}{
\begin{tabular}{l p{7.5cm} }
\toprule
 \multicolumn{2}{l}{\bf Visual input example:}  \\
\midrule
&  \includegraphics[height=5cm]{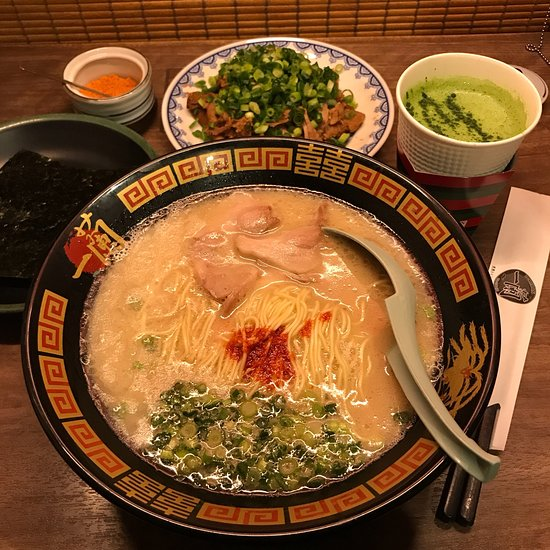} \\

User & If someone were to recommend a new flavor or topping to the dish, describe the reason for this change and how it might alter the overall taste. \\
\midrule
LLaVA-v1.5-7B & If someone were to recommend a new flavor or topping to the dish, they might suggest adding a spicy element, such as chili flakes or crushed red pepper, to enhance the taste and make it more flavorful. The current dish consists of a bowl of noodles with chicken, vegetables, and sauce, which is already a delicious and hearty meal. However, adding a spicy component could elevate the dish by providing a contrasting flavor and a kick of heat, making it more exciting and satisfying for those who enjoy spicy foods. This change would alter the overall taste by introducing a new dimension of flavor and complexity to the dish, making it more enjoyable and memorable for the person trying it. \\
\midrule
VARGPT & If someone were to recommend a new flavor or topping to the dish, it could be due to a desire to add a new taste dimension or to cater to personal preferences. For example, they might suggest adding a slice of avocado to the dish, which would introduce a creamy, buttery texture and a mild, nutty flavor that complements the richness of the noodles and the savory broth. This addition would alter the overall taste by providing a contrasting texture and a new layer of flavor complexity, enhancing the dish's overall appeal and making it more satisfying to eat.\\
\bottomrule
\end{tabular}
}
\captionof{table}{The case of visual understanding in VARGPT shows that our VARGPT has achieved better understanding performance.}
\label{tab:tricky_example-5}
  \end{minipage}
\end{table}

\newpage

{
    \small
    \bibliographystyle{ieeenat_fullname}
    \bibliography{main}
}


\end{document}